\documentclass{article}

\PassOptionsToPackage{authoryear, sort&compress}{natbib}

\usepackage[final]{neurips_2024}

\newif\ificml
\icmlfalse

\usepackage[colorinlistoftodos,prependcaption,textsize=tiny]{todonotes}
\usepackage{xargs} 
\usepackage{wrapfig}
\usepackage{threeparttable}
\usepackage{enumitem}
\usepackage{makecell}
\usepackage{tabularx}
\usepackage[utf8]{inputenc} %
\usepackage[T1]{fontenc}    %
\usepackage{CJKutf8}        %
\usepackage{url}            %
\usepackage{booktabs}       %
\usepackage{amsfonts}       %
\usepackage{nicefrac}       %
\usepackage{xcolor}         %
\usepackage{colortbl}       %
\usepackage{microtype}
\usepackage{graphicx}
\usepackage{pgfplots}
\usepgfplotslibrary{groupplots}
\pgfplotsset{compat=1.18}
\usepackage{floatrow}

\usepackage{subcaption}
\usepackage{newfloat}
\usepackage{xurl} 
\usepackage{listings}
\usepackage{fancyvrb}
\usepackage{longtable}
\usepackage{array}
\usepackage[breakable]{tcolorbox} %
\tcbuselibrary{skins}
\usepackage{multirow}
\usepackage{marvosym}
\usepackage{fontawesome5}  %

\definecolor{titleblue}{RGB}{25, 70, 135}
\definecolor{linkblue}{RGB}{50, 120, 200}
\definecolor{iconred}{RGB}{200, 60, 60}
\definecolor{iconpurple}{RGB}{120, 80, 160}
\definecolor{iconblue}{RGB}{60, 100, 180}
\definecolor{boxblue}{RGB}{230, 242, 255}  %

\newtcolorbox{titlebox}{
    colback=boxblue,
    colframe=boxblue,
    arc=4pt,
    boxrule=0pt,
    left=12pt,
    right=12pt,
    top=10pt,
    bottom=10pt,
    width=\textwidth,
    breakable,
}

\newcommand{\customtitle}[1]{%
    \begin{center}
    {\fontsize{18}{22}\selectfont\bfseries\color{titleblue}#1}
    \end{center}
    \vspace{0.3cm}
}

\definecolor{hzgrad}{RGB}{0, 150, 168}
\newcommand{\titleour}{%
    {\fontsize{19}{23}\selectfont\sffamily\bfseries
     \textcolor{titleblue!100!hzgrad}{H}%
     \textcolor{titleblue!90!hzgrad}{a}%
     \textcolor{titleblue!80!hzgrad}{r}%
     \textcolor{titleblue!70!hzgrad}{n}%
     \textcolor{titleblue!60!hzgrad}{e}%
     \textcolor{titleblue!50!hzgrad}{s}%
     \textcolor{titleblue!45!hzgrad}{s}%
     \textcolor{titleblue!40!hzgrad}{-}%
     \textcolor{titleblue!30!hzgrad}{Z}%
     \textcolor{titleblue!20!hzgrad}{e}%
     \textcolor{titleblue!10!hzgrad}{r}%
     \textcolor{titleblue!0!hzgrad}{o}}%
}

\newcommand{\corresp}{\Letter}        %

\newcommand{\customauthors}[1]{%
    \begin{center}
    #1
    \end{center}
    \vspace{0.1cm}
}

\newcommand{\customaffiliations}[1]{%
    \begin{center}
    \footnotesize
    #1
    \end{center}
}

\newenvironment{customabstract}{%
    \vspace{0.2cm}%
    \noindent\textbf{Abstract:}\par\vspace{0.35em}%
    \setlength{\parskip}{0.4em}%
    \setlength{\parindent}{0pt}%
}{%
    \par\vspace{0.55cm}%
}

\newcommand{\keywords}[1]{%
    \vspace{0.1cm}%
    \noindent\textbf{Keywords:}~~#1\par
    \vspace{0.4cm}%
}

\newcommand{\paperdate}[1]{%
    \noindent{\small\textcolor{iconred}{\faIcon{calendar-alt}}~\textbf{Date:}~#1}\par
}

\newcommand{\githubrepo}[1]{%
    \vspace{0.1cm}%
    \noindent{\small\textcolor{iconpurple}{\faIcon{github}}~\textbf{Github Repo:}~\textcolor{linkblue}{\url{#1}}}\par
}

\newcommand{\contactemail}[1]{%
    \vspace{0.1cm}%
    \noindent{\small\textcolor{iconblue}{\faIcon{envelope}}~\textbf{Contact:}~~#1}%
}

\usepackage{fancyhdr}

\AtBeginDocument{%
    \setlength{\headheight}{32pt}%
    \setlength{\headwidth}{\textwidth}%
}

\renewcommand{\headrulewidth}{0.4pt}

\fancypagestyle{firstpage}{%
    \fancyhf{}%
    \fancyhead[L]{%
        \includegraphics[height=22pt,keepaspectratio]{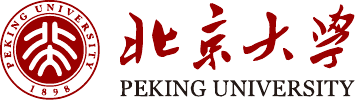}\hspace{14pt}%
        \raisebox{1pt}{\includegraphics[height=20pt]{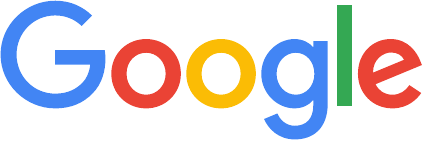}}\hspace{14pt}%
        \raisebox{1pt}{\includegraphics[height=20pt]{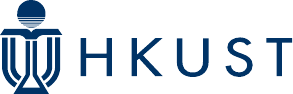}}%
    }%
    \fancyhead[R]{\our}%
    \renewcommand{\headrulewidth}{0.4pt}%
}

\usepackage[bookmarks=true,colorlinks=true,linkcolor=blue!60!black,urlcolor=magenta,citecolor=green!40!black]{hyperref}

\usepackage{amsmath}
\usepackage{amssymb}
\usepackage{mathtools}
\usepackage{amsthm}
\theoremstyle{plain}

\theoremstyle{definition}

\theoremstyle{remark}

\usepackage{bm} 
\usepackage{algorithm, algorithmic}

\usepackage{etoolbox}
\usepackage{natbib}
\newcounter{bibcount}
\makeatletter
\patchcmd{\@lbibitem}{\item[}{\item[\hfil\hspace{2.5em}\stepcounter{bibcount}{[\thebibcount]}\hspace{0.em}}{}{}
\makeatother
\usepackage[nameinlink,capitalise]{cleveref}

\usepackage{titletoc}
\crefname{section}{§}{§§}
\Crefname{section}{§}{§§}

\DeclareFloatingEnvironment[name=Prompt]{promptlisting}
\DeclareFloatingEnvironment[name=Skill]{skilllisting}
\DeclareFloatingEnvironment[name=Memory]{memorylisting}
\DeclareFloatingEnvironment[name=Code]{codelisting}

\crefname{lstlisting}{code}{code}
\Crefname{lstlisting}{Code}{Code}
\crefname{listing}{code}{code}
\Crefname{listing}{Code}{Code}
\crefname{promptlisting}{prompt}{prompts}
\Crefname{promptlisting}{Prompt}{Prompts}
\crefname{skilllisting}{skill}{skills}
\Crefname{skilllisting}{Skill}{Skills}
\crefname{memorylisting}{memory}{memories}
\Crefname{memorylisting}{Memory}{Memories}
\crefname{codelisting}{code}{code}
\Crefname{codelisting}{Code}{Code}

\usepackage{pifont}

\newcommand{\our}{\textsc{Harness-Zero}}
\newcommand{\term}[1]{\textbf{\textit{#1}}}  %

\definecolor{hzfull}{HTML}{3BA272}
\definecolor{hzheader}{RGB}{211,222,218}
\definecolor{hzalt}{RGB}{245,244,242}
\colorlet{hzgroup}{hzheader!45!white}
\definecolor{hzgain}{RGB}{79,124,101}
\definecolor{hzloss}{RGB}{168,108,102}
\newcommand{\gain}[1]{\,{\tiny\textcolor{hzgain}{$\uparrow$#1}}}
\newcommand{\loss}[1]{\,{\tiny\textcolor{hzloss}{$\downarrow$#1}}}
\newcommand{\nochange}[1]{\,{\tiny\textcolor{black!35}{$\uparrow$#1}}}
\newcommand{\snd}[1]{\underline{#1}}

\newcommand{\hztablesetup}{%
  \footnotesize
  \setlength{\tabcolsep}{5pt}%
  \renewcommand{\arraystretch}{1.15}%
}
\newcommand{\hztablerule}{\Xhline{1.1pt}}

\definecolor{hzlistingback}{RGB}{247,249,250}
\definecolor{hzlistingrule}{RGB}{176,188,194}
\definecolor{hzlistingcomment}{RGB}{42,112,75}
\lstdefinestyle{hzlisting}{
  basicstyle=\ttfamily\scriptsize,
  keywordstyle=\color{blue!55!black}\bfseries,
  commentstyle=\color{hzlistingcomment},
  stringstyle=\color{orange!55!black},
  backgroundcolor=\color{hzlistingback},
  frame=single,
  rulecolor=\color{hzlistingrule},
  breaklines=true,
  columns=fullflexible,
  keepspaces=true,
  showstringspaces=false,
  tabsize=2,
  xleftmargin=5pt,
  xrightmargin=5pt,
  framexleftmargin=4pt,
  framexrightmargin=4pt,
  aboveskip=7pt,
  belowskip=7pt,
  captionpos=b
}

\definecolor{hzboxframe}{RGB}{32,105,128}
\definecolor{hzboxback}{RGB}{242,248,250}
\newtcolorbox{hzkeybox}[1][\our{}]{
    enhanced,
    colback=hzboxback,
    colframe=hzboxframe,
    boxrule=0.7pt,
    arc=2mm,
    left=7pt,
    right=7pt,
    top=9pt,
    bottom=5pt,
    before skip=8pt,
    after skip=8pt,
    title=#1,
    fonttitle=\bfseries,
    coltitle=white,
    attach boxed title to top left={xshift=9pt,yshift=-2.5mm},
    boxed title style={
        colback=hzboxframe,
        colframe=hzboxframe,
        boxrule=0pt,
        arc=1.5mm,
        left=7pt,
        right=7pt,
        top=2pt,
        bottom=2pt
    }
}

\definecolor{softblue}{RGB}{100, 130, 230}    %
\definecolor{softcoral}{RGB}{200, 100, 110}   %
\definecolor{softred}{RGB}{200, 90, 90}       %
\definecolor{softgreen}{RGB}{90, 160, 110}    %

\definecolor{impgreen}{RGB}{60, 140, 90}  %

\definecolor{bestcolor}{RGB}{183, 223, 235}     %
\definecolor{secondcolor}{RGB}{225, 242, 248}   %
\definecolor{offlinecolor}{RGB}{232, 245, 233}  %
\definecolor{onlinecolor}{RGB}{255, 243, 224}   %

\begin{document}

\thispagestyle{firstpage}

\begin{titlebox}
\customtitle{\titleour:\\[0.35cm]Harness Distillation via Agent-as-Harness}

\customauthors{%
\textbf{Haoran Ye}\textsuperscript{1}\quad
\textbf{Yuxing Lu}\textsuperscript{2,3}\quad
\textbf{Haonan Dong}\textsuperscript{1}\quad
\textbf{Zhaochen Su}\textsuperscript{4}\quad
\textbf{Guojie Song}\textsuperscript{1,\,\corresp}
}

\customaffiliations{%
\textsuperscript{1}State Key Laboratory of General Artificial Intelligence,
School of Intelligence Science and Technology, Peking University\\[0.05cm]
\textsuperscript{2}College of Future Technology, Peking University\\[0.05cm]
\textsuperscript{3}Google\\[0.05cm]
\textsuperscript{4}The Hong Kong University of Science and Technology
}

\begin{customabstract}
Agent harnesses, the external systems that mediate model-environment interaction, can substantially improve agent performance, but their gains remain tied to the harness at deployment. Because the best harness varies across domains, instances, and models, a general-purpose agent must either settle for a suboptimal shared harness or route among an ever-growing set of specialized ones.
We therefore study \emph{agent harness distillation}: using a domain- or instance-optimized harness as training-time guidance and transferring the behaviors it induces into model weights, so that its gains survive under a single fixed target harness.
The challenge is that the two harnesses differ in action space and available information, so guidance from the optimized harness cannot serve directly as supervision for the target one.
We introduce \our{}, which enables harness distillation through \emph{agent-as-harness}. Guided by the optimized harness, a \emph{harnessing agent} corrects student responses before execution in the target harness's action space, turning harness guidance into training demonstrations.
Fine-tuning on the resulting trajectories internalizes harness-induced behavior into the model, so the specialized harness can be removed at deployment.
Our experiments spanning knowledge work, tool use, and science domains show that:
\ding{182} For frontier LLMs using the same evolved harness, agent-as-harness outperforms code-as-harness.
\ding{183} With the specialized harness removed at deployment, \our{} improves the base model's macro-average task success from 23.3\% to 44.3\%, even exceeding the 41.7\% it reaches with that harness still attached, while reducing inference token usage by 27\%--86\%.
\ding{184} \our{} recovers harness-induced behaviors absent from the base model, with 82.3\% average recovery across 28 patterns in the three domains.

\end{customabstract}


\noindent
\begin{minipage}[t]{0.75\textwidth}
\paperdate{September 2026}
\githubrepo{https://github.com/metaevo-ai/harness-zero}
\contactemail{\href{mailto:hrye@stu.pku.edu.cn}{\textcolor{linkblue}{hrye@stu.pku.edu.cn}} \quad \href{mailto:gjsong@pku.edu.cn}{\textcolor{linkblue}{gjsong@pku.edu.cn}}}
\end{minipage}%
\hfill
\end{titlebox}

\vspace{0.3cm}

\section{Introduction}\label{sec:introduction}

An LLM agent's capabilities depend on both its model and its \term{harness}: the external system that organizes tool use, manages context and state, and controls interaction with the environment~\citep{weng2026harness,ning2026codeharness}. Harness engineering has become a central lever for improving agent performance. 
Coding-agent harnesses combine shell access, file systems for persistent memory, subagents, and background jobs~\citep{weng2026harness,yang2024swe}. Research-agent harnesses organize workflows for hypothesis generation, experimentation, and evidence collection~\citep{lu2026aiscientist}. Context management and experience reuse further support continual learning and long-horizon execution~\citep{ye2026mce,ma2026longhorizon,karten2026continual,karten2026prime,yan2026harnessofharness,ye2024reevo}. 
Recent methods such as Meta-Harness automate this engineering process by optimizing harness code~\citep{lee2026metaharness,lin2026ahe,zhang2026selfharness}.

Harness optimization, however, improves the agent's external scaffolding rather than the model itself, so its gains remain tied to that harness at deployment. Because the best harness varies across domains, instances, and base models~\citep{zhang2026selfharness,luo2026harnessbank,liu2026interference}, a general-purpose agent must choose between a shared harness and a collection of specialized ones. A shared harness forgoes some specialized gains~\citep{luo2026harnessbank,liu2026interference,yao2026harnessbench}. Maintaining many harnesses instead requires routing and incurs recurring costs in context, model calls, tool calls, and orchestration~\citep{zhang2025agentic,lin2026provisioning,chang2026agentsysbench,zhang2025cut}.
Neither choice moves the discovered harness improvements into the model.

\begin{hzkeybox}
To address these limitations, we introduce \textbf{\our{}}, a framework for \term{agent harness distillation}. It uses an optimized harness as training-time guidance to distill its induced behaviors into model parameters. This turns gains discovered through harness optimization into model capabilities available under a fixed target harness.
\end{hzkeybox}

The distillation target in \our{} spans specialized tool use expressed in the student's native action space, behavioral patterns enforced by middleware code, and accumulated knowledge and reusable experience supplied by skills and memory.
The source and target harnesses can differ in both action space and available information, making trajectories collected under the source harness unsuitable for direct imitation under the target harness.

Our solution to this challenge is \term{agent-as-harness}, which wraps the student agent with a \term{harnessing agent} at its response boundary.
We denote the student's fixed operating harness as the target harness $h$, the evolved student-side harness (via, e.g., meta-harness \citep{lee2026metaharness}) as $h^\star$, and its adaptation for the harnessing agent as the private reference harness $\mathcal K$.
As one example of this adaptation, a middleware rule in $h^\star$ that blocks risky actions becomes a review-time warning in $\mathcal K$, activated when the student proposes such an action.

During data collection, the harnessing agent uses $\mathcal K$ to review each proposed response before it is executed or added to the trajectory. It passes sound proposals unchanged and otherwise makes the smallest coherent correction, expressed as a complete response in the student's action space. The accepted response is executed through $h$, and the resulting observation is appended to the student's trajectory. The harnessing agent cannot inspect the student sandbox or consult hidden task solutions, and its private review discussion remains outside the student-visible trajectory.
\our{} applies supervised fine-tuning (SFT) to these reviewed rollouts to internalize the demonstrated behaviors in model parameters.

\begin{hzkeybox}[Agent-as-harness]
Agent-as-harness builds an adaptive mapping between harnesses: an agent reads the guidance of one harness and re-expresses it, response by response, as executable corrections in the action space of another. Any optimized harness can therefore supply training demonstrations for the target harness, regardless of how the two differ in actions or available information.
\end{hzkeybox}

We evaluate \our{} across three domains: spreadsheet-based knowledge work (SpreadsheetBench Verified), multi-application tool use (AppWorld), and scientific reasoning (USPTO Retrosynthesis). We instantiate $h$ as a fixed, minimal mini-SWE-agent~\citep{sweagentteam2025miniswe} with one Bash \texttt{execute} tool. We first evaluate agent-as-harness on frontier models without training and find that it outperforms code-as-harness (81.1\% vs.\ 78.1\%, averaged across six benchmark--model settings). We then evaluate harness distillation on Qwen3.5-9B and remove $h^\star$, $\mathcal K$, and the harnessing agent at deployment. Under $h$ alone, \our{} raises macro-average performance from 23.3\% to 44.3\%, exceeding the 41.7\% obtained by the base model with $h^\star$ still attached, while cutting per-task inference tokens by 27\%--86\%. Moreover, a single jointly trained model can consolidate all three domain harnesses without inference-time routing.
Controlled ablations further show that harness-guided review produces substantially better distilled performance than alternative supervision sources, including direct trajectories from a stronger model, trajectories generated under $h^\star$, review without $\mathcal K$, and review given only the task answer (30\% vs.\ 3--15\%). Behavioral analysis also finds that the distilled model recovers most behaviors induced by $h^\star$ but absent from the base model (82.3\% recovery on average across 28 patterns).

\noindent\textbf{Contributions.}
\ding{182} We formulate \emph{agent harness distillation} and present \our{}, which transfers behaviors induced by an optimized harness into model parameters for deployment under a fixed target harness.
\ding{183} We introduce \emph{agent-as-harness}, which translates guidance from an optimized harness into executable supervision at the student's response boundary, enabling imitation learning across different harness action spaces.
\ding{184} Experimental results show that agent-as-harness can outperform code-as-harness, that \our{} retains the gains of optimized harnesses after they are removed, and that the distilled model recovers harness-induced behaviors.

\section{Related work}\label{sec:related}

\paragraph{Code-as-harness and its optimization.}
Harnesses mediate the interaction between LLMs and their environments through tools, context management, control flow, and persistent state~\citep{weng2026harness,ning2026codeharness,zhou2026externalization}.
Modern coding agents such as Claude Code~\citep{anthropic2026claudecode}, Codex~\citep{openai2026codex}, Kimi Code~\citep{moonshot2026kimicode}, Pi~\citep{earendil2026pi}, and OpenCode~\citep{anomaly2026opencode} embody different design philosophies within this space.
Natural-Language Agent Harnesses represent run-level policies as editable documents, which an agent interprets into actions~\citep{pan2026natural}.
Automatic, training-free optimization has expanded from prompts~\citep{khattab2023dspy,fernando2023promptbreeder,agrawal2025gepa} and context~\citep{zhang2025agentic,ye2026mce} to workflows~\citep{hu2025adas,zhang2025aflow} and harnesses~\citep{lee2026metaharness,lin2026ahe,zhang2026selfharness}.
\citet{jitagent2026} train a model to generate per-task harnesses just in time.
Harness design can also support test-time strong-to-weak transfer, in which a stronger model builds an inference-time harness for a fixed weaker model~\citep{qian2026ai4ai}.
Complementing this line of work, \our{} introduces agent-as-harness, which can outperform code-as-harness for frontier models and supports distilling optimized harness behavior into model weights.

\paragraph{Co-evolution of model and harness.}
Several approaches combine harness optimization with parameter updates.
One line alternates between the two, using revised harnesses to generate data for the next model update and updated models to motivate further harness search~\citep{hebbar2026sia,chen2026coharness,kim2026whale,karten2026continual}.
Other methods optimize the two jointly, treating model--harness compatibility as the objective and adapting the agent harness together with the policy trained from its trajectories~\citep{chen2026harnessforge,chen2026harnessx,luo2026hase,mao2026safeevolve}.
These studies show that harness and weight updates can reinforce each other, but the resulting gains may remain coupled to the harness.
A controlled coding-agent study supports this concern~\citep{le2026multiharness}: changing the evaluation harness affected performance more than the training method, and training with feedback collected across harnesses did not improve transfer to a held-out minimal ReAct harness.
\our{} instead uses optimized harnesses to guide a temporary harnessing agent and distills the resulting behavior into model weights, internalizing harness gains without retaining or routing the harness collection at deployment.

\paragraph{Distillation from privileged guidance.}
Neural knowledge distillation seeks to compress the behavior and capabilities of a teacher system into a student network~\citep{schmidhuber1991chunker,schmidhuber1992chunker,hinton2015distilling}.
In the context of LLM agents, EvoHarness-RL provides early evidence of harness internalization on ALFWorld, where its trained agent learns to manage external state and makes fewer, more selective harness calls~\citep{ning2026evoharness}.
OPHSD distills privileged outputs produced by sequential draft--verify and plan--solve LLM workflows into a standalone model~\citep{zhao2026ophsd}.
Together, these studies provide preliminary evidence that models can absorb harness-induced behavior into their parameters.
However, each addresses only an individual mechanism rather than a complete tool-using agent harness; EvoHarness-RL retains the external workspace at deployment, and OPHSD does not involve an interactive agent loop.
\our{} introduces a general method for distilling complete tool-using agent harnesses into a model that runs under a minimal target harness at deployment.

\section{\our{}}\label{sec:method}

\begin{figure}[t]
\centering
\includegraphics[width=\linewidth]{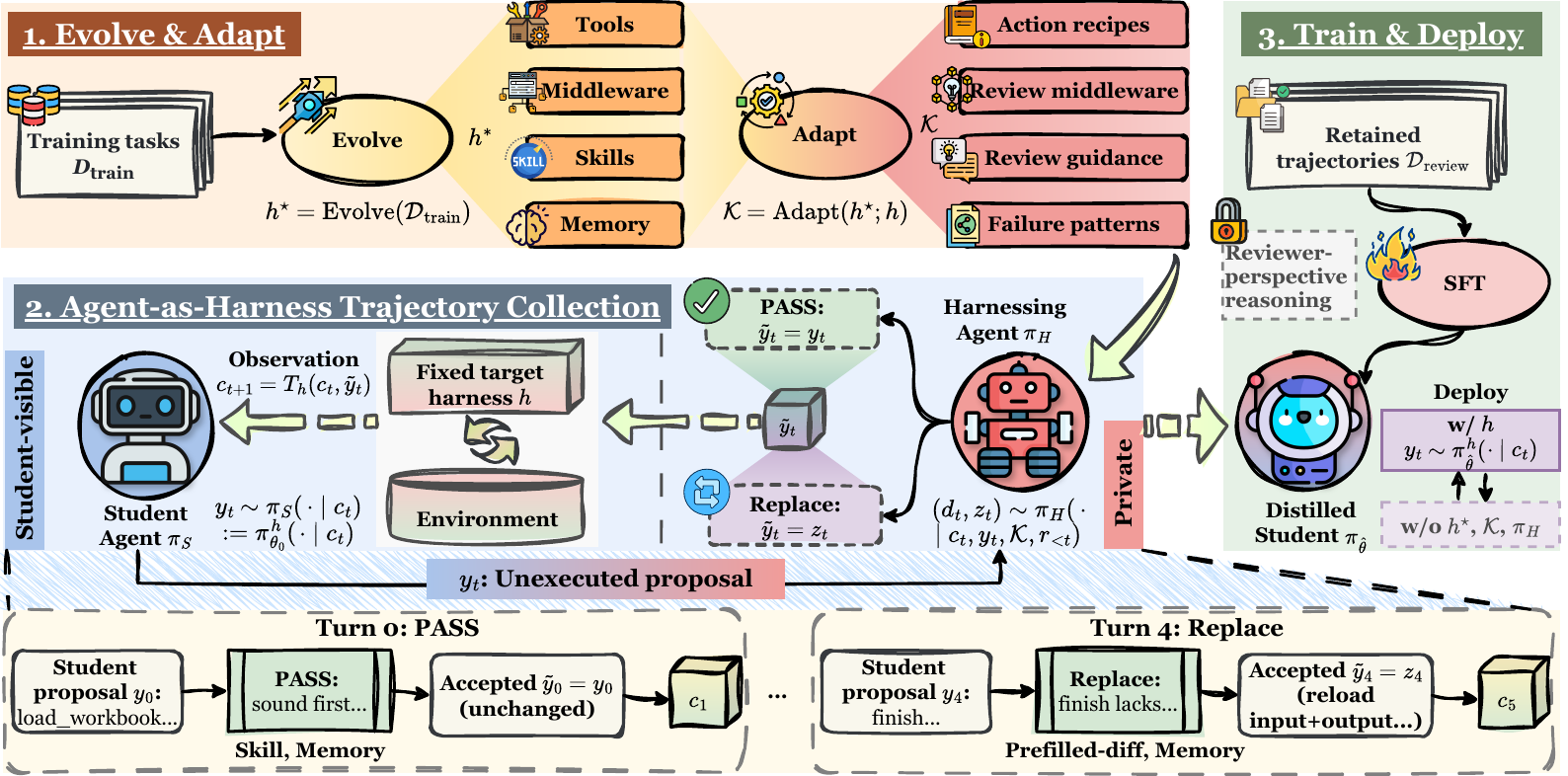}
\caption{Overview of \our{}. \textbf{(1) Evolve and adapt.} We evolve a student-side harness $h^\star$ on training tasks, and adapt it into a private reference harness $\mathcal K$ relative to the fixed target harness $h$. \textbf{(2) Agent-as-harness trajectory collection.} The student $\pi_S$ runs under $h$ and proposes response $y_t$ at each turn. The harnessing agent $\pi_H$ reviews it using $\mathcal K$ and corrects it ($\tilde y_t$) when needed. $\tilde y_t$ is executed through $h$ and enters the student's context, while the review remains private. \textbf{(3) Train and deploy.} We apply SFT to the reviewed trajectories $\mathcal D_{\mathrm{review}}$, masking any inadvertent reviewer-perspective reasoning from the loss. The distilled student is deployed under $h$ alone.}
\label{fig:teaser}
\end{figure}

\our{} trains a student model to reproduce behaviors induced by an evolved harness.
The method has three stages (\Cref{fig:teaser}). First (\Cref{sec:method-translation}), we evolve a student-side harness on training tasks and adapt it into a private reference harness for a separate \emph{harnessing agent}. Second (\Cref{sec:method-aah}), the harnessing agent wraps the student's response loop for training trajectory collection. Third (\Cref{sec:method-training}), we apply SFT to reviewed trajectories jointly produced by the student and the harnessing agent.
At deployment, the distilled student retains the evolved harness's gains under the target harness alone.

\vspace{-13pt}

\subsection{Harness evolution and adaptation}\label{sec:method-translation}

The first stage builds the domain-specific guidance used during training trajectory collection. We denote the fixed target harness by $h$, the evolved student-side harness by $h^\star$, and the private reference harness adapted from it by $\mathcal K$. Given training tasks $\mathcal D_{\mathrm{train}}$, we write the two steps as
\begin{equation}
h^\star=\operatorname{Evolve}(\mathcal D_{\mathrm{train}}),
\qquad
\mathcal K=\operatorname{Adapt}(h^\star;h).
\label{eq:evolve-adapt}
\end{equation}

\paragraph{Harness evolution.}
We evolve $h^\star$ on training tasks. This step can use existing automated harness-optimization methods~\citep{lee2026metaharness,lin2026ahe,zhang2026selfharness}. In our implementation, a simple skill-guided evolution loop analyzes recurring failures and updates the domain harness. The resulting $h^\star$ follows the DeepAgents abstraction~\citep{langchain_deepagents}, which includes tools, middleware, skills, and memory.
\Cref{app:harness-evolution} describes the three-round procedure and presents the evolution skill.

\paragraph{Harness adaptation.}
The evolved student-side harness $h^\star$ is designed to act directly around the student. Tools extend its action space, middleware modifies or blocks its execution loop, and skills and memory instruct the student. We adapt these components into the reference harness $\mathcal K$ used by the harnessing agent; the process can be automated by an agent.
The adaptation is relative to $h$, because any correction the harnessing agent constructs from $\mathcal K$ must ultimately be executed by the student in the target harness's action space.
In general, the adaptation preserves the harness components' intended behavior while changing their audience and enforcement point. Tools become specifications for constructing student-native equivalents; student-side middleware becomes review middleware that privately alerts the harnessing agent when the corresponding condition is met; and skills and memory become diagnostic criteria and intervention guidance. The adaptation may also produce a domain prompt appended to the harnessing agent's system prompt, stating the domain's review policy.
\Cref{app:harness-adaptation} gives detailed examples of such adaptation.

\subsection{Agent-as-harness}\label{sec:method-aah}

Let $\pi_\theta^h$ denote the response policy induced when a model with parameters $\theta$ operates under $h$, and let $T_h(c,y)$ denote the transition function that executes response $y$ through $h$ from context $c$ and returns the next student-visible context.

\textbf{Code-as-harness}, after harness evolution, runs the base student with parameters $\theta_0$ directly under $h^\star$~\citep{ning2026codeharness}:
\begin{equation}
y_t\sim\pi_{\theta_0}^{h^\star}
    (\cdot\mid c_t),
\qquad
c_{t+1}=T_{h^\star}(c_t,y_t).
\label{eq:code-harness}
\end{equation}

\textbf{Agent-as-harness} instead runs the student under $h$, with a harnessing agent wrapping it at its response boundary.
The harnessing agent intercepts each proposed response before execution and either passes it or replaces it. It uses $\mathcal K$ for private guidance on when to intervene and how to construct a replacement; $\mathcal K$ neither acts on the environment nor enters the student-visible context. The target harness $h$ defines the student's action space and executes the accepted response. The student and harnessing agent may use the same underlying model, with different instructions and context for their respective roles.

At turn $t$, let $c_t$ denote the student's visible context: the task, previous accepted responses, and resulting environment observations. The student policy $\pi_S$ proposes a response $y_t$, which the harnessing policy $\pi_H$ reviews using $c_t$, the guidance in $\mathcal K$, and its private history $r_{<t}$:
\begin{align}
y_t &\sim \pi_S(\,\cdot\mid c_t)
      :=\pi_{\theta_0}^{h}(\,\cdot\mid c_t), \nonumber\\
(d_t,z_t) &\sim \pi_H(\,\cdot\mid c_t,y_t,\mathcal K,r_{<t}), \nonumber\\
\tilde y_t &=
\begin{cases}
y_t, & d_t=\textsc{pass},\\
z_t, & d_t=\textsc{replace},
\end{cases}\nonumber\\
c_{t+1} &= T_h(c_t,\tilde y_t),
\label{eq:review}
\end{align}
where $d_t$ is the review decision, $z_t$ is a complete replacement response valid under $h$, and $r_{<t}$ contains earlier review exchanges and the harnessing agent's prior file-system reads from $\mathcal K$. A single harnessing-agent session spans the entire student rollout. At each review, it receives the student-visible events added since the previous review and the current unexecuted proposal. Only the accepted response $\tilde y_t$ enters the student-visible trajectory. Any actions it contains are then executed through $h$, and their observations become part of $c_{t+1}$. The rejected proposal and private review remain outside this trajectory.
This process realizes source-harness guidance as a target-harness trajectory incrementally, with each correction conditioned on the student's current interaction history.
\Cref{app:review-protocol} specifies the review process and gives the harnessing agent's system prompt.

\begin{hzkeybox}[Code-as-harness vs.\ agent-as-harness]
\textbf{Code-as-harness} (\cref{eq:code-harness}) runs the student directly under the evolved $h^\star$, which shapes and executes $y_t$.

\smallskip
\textbf{Agent-as-harness} (\cref{eq:review}) runs the student under $h$, while a harnessing agent reviews the student's proposal $y_t$ using $\mathcal K$, whose guidance is adapted from $h^\star$. The accepted response $\tilde y_t$ is then executed through $h$.
\end{hzkeybox}

\paragraph{Intervention policy and constraints.}
Guidance from $\mathcal K$ steers rollouts toward behaviors and states that an unaided student under $h$ may not reach. To facilitate SFT, the harnessing agent minimizes changes to the student's proposals. It passes sound proposals unchanged. When intervention is necessary, it makes the smallest coherent correction needed to follow $\mathcal K$'s guidance and preserves the rest of the proposal whenever possible. Each replacement is a complete response valid under $h$ that continues from the current student-visible state.

The harnessing agent's privileged access is limited to reading $\mathcal K$. It cannot inspect hidden solutions or verifier feedback, nor can it access environment state outside $c_t$. Any additional task evidence must therefore be obtained by proposing an action available under $h$. For example, it can replace a premature completion with code that checks the student's work. Executing the code through $h$ adds both the check and its result to the student-visible trajectory, grounding the intervention in student-observable evidence. Together, these intervention and grounding constraints make the reviewed trajectories directly usable to fine-tune the student for operation under $h$.

\subsection{Learning from reviewed trajectories}\label{sec:method-training}

We perform imitation learning under the target harness by applying SFT to the accepted responses in reviewed rollouts (\cref{eq:review}), including both unchanged student proposals and harness-guided replacements. Replacements should be self-contained responses written from the student's perspective. Because they are generated within the private review context, they may inadvertently include reviewer-perspective reasoning about the student's proposal or the review decision. We mask such reasoning from the loss; \Cref{app:data-collection} details data collection and the filtering rule. Let $\mathcal D_{\mathrm{review}}$ denote the retained trajectories, we optimize:
\begin{equation}
\hat{\theta}
=\arg\min_\theta
    -\sum_{\tau\in\mathcal D_{\mathrm{review}}}
     \sum_{t=1}^{T_\tau}
    \log \pi_\theta^{h}(\tilde y_t\mid c_t).
\label{eq:sft}
\end{equation}
Here, $T_\tau$ is the number of accepted response turns in trajectory $\tau$. At deployment, the distilled policy acts directly under the same target harness:
\begin{equation}
y_t\sim\pi_{\hat{\theta}}^{h}(\cdot\mid c_t),
\qquad
c_{t+1}=T_h(c_t,y_t).
\label{eq:deployment}
\end{equation}
The deployed system is therefore $(\pi_{\hat{\theta}},h)$, without $h^\star$, $\mathcal K$, or the harnessing agent. The training objective is for $\pi_{\hat{\theta}}$ to reproduce under $h$ the behavior patterns induced by $h^\star$.

\section{Experiments}\label{sec:experiments}

We first compare agent-as-harness with code-as-harness at inference time, then test whether \our{} can distill an optimized harness into model weights.

\subsection{Experimental setup}\label{sec:exp-setup}

\paragraph{Tasks and splits.}
We evaluate three task domains.
(1) \textbf{SpreadsheetBench Verified} contains 400 real-world spreadsheet-manipulation tasks~\citep{ma2024spreadsheetbench}; we use 300 for harness evolution and training data collection and hold out 100 for evaluation.
(2) \textbf{AppWorld} evaluates interactive tool use across simulated applications~\citep{trivedi2024appworld}; we merge its official train and development sets into a 147-task training split and hold out the 168 \texttt{test\_normal} tasks, grouped into 56 three-task scenarios.
(3) \textbf{USPTO Retrosynthesis} covers single-step precursor prediction from the USPTO reaction corpus~\citep{lowe2012extraction,jin2017predicting}; we use a 500-task training split balanced across its ten reaction classes and a disjoint 100-task test split.
All three run in the Harbor framework~\citep{harbor2026,shi2026harbor}.
We report single-run task success (pass@1, \%) for SpreadsheetBench and USPTO and scenario goal completion (SGC, \%) for AppWorld.

\paragraph{Models and training.}
The target harness $h$ is a minimal mini-SWE-agent-style harness~\citep{sweagentteam2025miniswe} with a fixed system prompt and a single Bash execution tool.
The training-free experiments use GPT-5.6 Sol~\citep{openai2026gpt56} and DeepSeek-V4-Pro~\citep{deepseekai2026deepseekv4}. The distillation experiments use Qwen3.5-9B~\citep{qwen2026qwen35} as the base model and GPT-5.6 Sol as the harnessing agent; harness evolution uses Kimi K3~\citep{moonshot2026kimik3} under Kimi Code. Reasoning is enabled for all models, with reasoning effort set to \emph{high} when applicable.
After rollout collection and filtering, the training data comprise 487 rollouts for SpreadsheetBench, 282 for AppWorld, and 500 for USPTO.
We perform LoRA SFT on Qwen3.5-9B for two epochs using the Tinker recipe~\citep{thinkingmachines2025tinker}.
\Cref{app:implementation} gives the student prompt, the model access routes, and the full training configuration.

\subsection{Evaluating agent-as-harness}\label{sec:exp-where}

We first compare agent-as-harness against code-as-harness at inference time, with no parameter updates.
\Cref{tab:train-free} varies two factors: whether the evolved harness is available, and whether it reaches the student as code wrapped around it or as a harnessing agent reviewing its responses.
The two code-as-harness conditions use no harnessing agent: mini-SWE-agent runs the student under $h$ alone, and meta-harness mounts $h^\star$ on top of it.
The two agent-as-harness conditions keep the student under $h$ and add a harnessing agent that consults either an empty reference harness (w/o evolved) or $\mathcal K$ adapted from $h^\star$ (w/ evolved).
In both, the same model plays student and harnessing agent, so the gains cannot come from a stronger supervising model.
On the frontier models we reuse the $h^\star$ evolved on Qwen3.5-9B.

\begin{table}[!t]
\centering
\footnotesize
\setlength{\tabcolsep}{4pt}
\renewcommand{\arraystretch}{1.05}
\caption{Comparing agent-as-harness against code-as-harness, with and without the evolved harness. Best result per benchmark--model setting in bold, second best underlined.}
\label{tab:train-free}
\begin{tabular}{llcccc}
\Xhline{1.1pt}
\rowcolor{hzheader}
\textbf{Benchmark} & \textbf{Model} & \multicolumn{2}{c}{\textbf{Code-as-Harness}} & \multicolumn{2}{c}{\textbf{Agent-as-Harness}} \\
\rowcolor{hzheader}
& & \makecell{mini-SWE-agent\\($h$)} & \makecell{meta-harness\\($h^\star$)} & \makecell{w/o evolved\\($h,\varnothing$)} & \makecell{w/ evolved\\($h,\mathcal K$)} \\
\Xhline{1.1pt}
SpreadsheetBench & DeepSeek-V4-Pro & 76.0 & 77.0\gain{1.0} & \snd{78.0}\gain{2.0} & \textbf{86.0}\gain{10.0} \\
& GPT-5.6 Sol & 83.0 & \snd{84.0}\gain{1.0} & 83.0\nochange{0.0} & \textbf{88.0}\gain{5.0} \\
\hline
AppWorld & DeepSeek-V4-Pro & 75.0 & \textbf{91.1}\gain{16.1} & 76.8\gain{1.8} & \snd{89.3}\gain{14.3} \\
& GPT-5.6 Sol & \textbf{96.4} & \snd{94.6}\loss{1.8} & \snd{94.6}\loss{1.8} & \textbf{96.4}\nochange{0.0} \\
\hline
USPTO & DeepSeek-V4-Pro & 31.0 & \snd{56.0}\gain{25.0} & 30.0\loss{1.0} & \textbf{60.0}\gain{29.0} \\
& GPT-5.6 Sol & 50.0 & \snd{66.0}\gain{16.0} & 53.0\gain{3.0} & \textbf{67.0}\gain{17.0} \\
\Xhline{1.1pt}
\multicolumn{2}{l}{\textbf{Average}} & 68.6 & \snd{78.1}\gain{9.5} & 69.2\gain{0.6} & \textbf{81.1}\gain{12.5} \\
\multicolumn{2}{l}{\textbf{Relative improvement over $h$}} & --- & \snd{+22.5\%} & +1.0\% & \textbf{+27.6\%} \\
\Xhline{1.1pt}
\end{tabular}

\end{table}

\noindent\textbf{Obs.\ding{182} With evolved harness, agent-as-harness outperforms code-as-harness on average.}
Across the six settings in \Cref{tab:train-free}, agent-as-harness with the adapted $\mathcal K$ averages 81.1\%, against 78.1\% for meta-harness and 68.6\% for mini-SWE-agent.
With an empty $\mathcal K$ it averages only 69.2\%, so review alone explains little of the gain.
Beyond these benchmark numbers, agent-as-harness also offers better adaptability across model updates. 
A code harness encodes assumptions about how a model should act, and as capabilities change those assumptions go stale, forcing the harness to be re-adapted for each new model~\citep{qian2026ai4ai,liu2026interference}.
Agent-as-harness moves that adaptation into inference: the harnessing agent interprets $\mathcal K$ against the current trajectory and decides when and how to intervene, so the guidance stays reusable and a stronger harnessing model directly improves how it is applied.
While this approach requires a sufficiently capable harnessing agent (\Cref{app:aah-analysis}), we expect its advantage over fixed code harnesses to widen as foundation models continue to improve.

\begin{table}[!t]
\centering
\footnotesize
\setlength{\tabcolsep}{5pt}
\renewcommand{\arraystretch}{1.05}
\caption{Harness distillation results across three benchmarks.}
\label{tab:distillation}
\begin{tabular}{lcccc}
\Xhline{1.1pt}
\rowcolor{hzheader}
\textbf{Setting} & \multicolumn{3}{c}{\textbf{Benchmark}} & \\[-2pt]
\rowcolor{hzheader}
& SpreadsheetBench & AppWorld & USPTO & \raisebox{1.1ex}[0pt][0pt]{\textbf{Avg.}} \\
\Xhline{1.1pt}
\rowcolor{hzgroup}
\multicolumn{5}{c}{\textit{Base model}} \\
mini-SWE-agent ($h$)     & 31.0 & 26.8 & 12.0 & 23.3 \\
meta-harness ($h^\star$) & 39.0\gain{8.0} & \snd{48.2}\gain{21.4} & \textbf{38.0}\gain{26.0} & \snd{41.7}\gain{18.4} \\
DeepAgents               & 35.0\gain{4.0} & 19.6\loss{7.2} & 7.0\loss{5.0} & 20.5\loss{2.8} \\
Claude Code              & 31.0\nochange{0.0} & 10.7\loss{16.1} & 6.0\loss{6.0} & 15.9\loss{7.4} \\
\rowcolor{hzgroup}
\multicolumn{5}{c}{\textit{Distilled model}} \\
\textbf{\our{} ($h$, multi-domain joint)} & \textbf{47.0}\gain{16.0} & 39.3\gain{12.5} & 21.0\gain{9.0} & 35.8\gain{12.5} \\
\textbf{\our{} ($h$, per-domain)} & \snd{44.0}\gain{13.0} & \textbf{58.9}\gain{32.1} & \snd{30.0}\gain{18.0} & \textbf{44.3}\gain{21.0} \\
\Xhline{1.1pt}
\end{tabular}

\end{table}

\begin{table}[!t]
\centering
\footnotesize
\setlength{\tabcolsep}{6pt}
\renewcommand{\arraystretch}{1.08}
\caption{Deployment token usage comparing the base model with $h^\star$ attached against \our{} deployed under the minimal harness $h$.}
\label{tab:token-efficiency}
\begin{tabular}{lcccc}
\Xhline{1.1pt}
\rowcolor{hzheader}
\textbf{Benchmark} & \multicolumn{2}{c}{\textbf{Avg.\ input tokens / task}} & \multicolumn{2}{c}{\textbf{Avg.\ output tokens / task}} \\
\rowcolor{hzheader}
& Base ($h^\star$) & \our{} ($h$) & Base ($h^\star$) & \our{} ($h$) \\
\Xhline{1.1pt}
SpreadsheetBench & 516.0K & \textbf{222.1K}\,{\tiny\textcolor{hzgain}{$\downarrow$56.9\%}} & 17.4K & \textbf{9.8K}\,{\tiny\textcolor{hzgain}{$\downarrow$43.7\%}} \\
AppWorld         & 339.8K & \textbf{249.5K}\,{\tiny\textcolor{hzgain}{$\downarrow$26.6\%}} & 5.0K  & \textbf{3.5K}\,{\tiny\textcolor{hzgain}{$\downarrow$28.8\%}} \\
USPTO            & 118.1K & \textbf{15.4K}\,{\tiny\textcolor{hzgain}{$\downarrow$87.0\%}}  & 9.5K  & \textbf{2.3K}\,{\tiny\textcolor{hzgain}{$\downarrow$75.8\%}} \\
\Xhline{1.1pt}
\end{tabular}
\end{table}

\subsection{Evaluating agent harness distillation}\label{sec:exp-distillation}

We next test whether the behavior induced by the optimized harness $h^\star$ can be retained after distillation, when $h^\star$, reference harness $\mathcal K$, and harnessing agent are removed.
\Cref{tab:distillation} compares the base model under $h$, the base model with $h^\star$ mounted, and the distilled model under $h$ (both per-domain specialists and a single multi-domain model trained on the union of all 1,269 trajectories).
As reference points, we also run the base model under two general-purpose harnesses: DeepAgents~\citep{langchain_deepagents}, the abstraction on which $h^\star$ is built (\Cref{sec:method-translation}), and Claude Code~\citep{anthropic2026claudecode}.

\noindent\textbf{Obs.\ding{183} Distillation raises the base model's macro-average performance by 21.0 points and surpasses $h^\star$.}
\our{} raises the macro average from 23.3\% to 44.3\%, a 21.0-point absolute gain and a 90.1\% relative improvement.
It also exceeds the 41.7\% macro average of the base model equipped with $h^\star$.
Neither general-purpose harness benefits the base model. On one hand, a 9B model handles their larger, generic tool suites and extended context poorly. On the other hand, domains such as USPTO and AppWorld demand domain-specific tooling and constraints present in $h^\star$ (such as molecular validation tools for USPTO), rendering generic tools beyond basic bash largely unhelpful and distracting.

\noindent\textbf{Obs.\ding{184} Procedural harness behavior is easier to internalize than deep domain knowledge.}
On SpreadsheetBench and AppWorld, \our{} under $h$ alone surpasses $h^\star$.
Their harnesses mainly encode recurring procedures for state inspection, targeted changes, and verification, which reviewed trajectories can demonstrate directly.
On USPTO, \our{} improves over the base model (30.0\% vs.\ 12.0\%) but trails $h^\star$ (38.0\%).
There, $h^\star$ also supplies reaction priors, candidate-generation logic, and executable SMILES validation; transferring this knowledge and functionality may require broader pretraining or mid-training coverage, or more distillation trajectories.

\noindent\textbf{Obs.\ding{185} \our{} cuts deployment token usage by 27\%--86\%.}
Retaining a specialized harness $h^\star$ at deployment forces every query to repeatedly pay for long skill and memory prompts, custom tool schemas, and middleware guard retries across turns.
As shown in \Cref{tab:token-efficiency}, internalizing $h^\star$ into the model under the minimal harness $h$ reduces average per-task input tokens by 56.9\% on SpreadsheetBench, 26.6\% on AppWorld, and 87.0\% on USPTO, and reduces output tokens by 43.7\%, 28.8\%, and 75.8\%, respectively.

\noindent\textbf{Obs.\ding{186} Multi-domain joint distillation consolidates incompatible harnesses into a single model without routing.}
Code harnesses from different domains cannot be trivially merged into a single runtime harness. Combining their skills, memories, tools, and middleware bloats the context window, introduces conflicting guard rules, and cross-component interference, or forces a general-purpose agent to maintain a router across domain harnesses.
Because \our{} translates every domain harness into trajectories in the shared action space of $h$, trajectories from distinct harnesses can be pooled directly.
Training a single model on the union of all three training sets (487 SpreadsheetBench, 282 AppWorld, and 500 USPTO trajectories) improves over the base model on every benchmark (+16.0 points on SpreadsheetBench, +12.5 on AppWorld, and +9.0 on USPTO; \Cref{tab:distillation}) and even surpasses the single-domain specialist on SpreadsheetBench (47.0\% vs.\ 44.0\%) without any inference-time routing.
We discuss that closing the remaining gap to single-domain specialists will benefit from scaling up data collection and tuning the domain mixture, as in standard general-purpose post-training.

\subsection{Comparing alternative distillation signals}\label{sec:exp-supervision}
To ablate the agent-as-harness recipe and understand what makes it effective, we first compare against alternative trajectory sources on USPTO (\Cref{tab:trajectory-source-ablation}), where all 500 collected trials per condition are kept without reward filtering.
We vary the rollout generator, the use of $h^\star$, and the private context available to the harnessing agent.
The three direct-distillation baselines collect rollouts from the teacher (GPT-5.6 Sol, used in the harnessing agent of \our{}) under $h$, the teacher with $h^\star$ mounted, and the base student with $h^\star$ mounted.
The two harnessing-agent controls keep the base student under $h$ and give the harnessing agent either an empty $\mathcal K$ or the oracle answer.

All conditions start from Qwen3.5-9B, use the same 500-task collection split and training recipe, and evaluate the 2-epoch checkpoint under $h$ on the test set. Trajectories are retained after the same structural and privacy checks.
For each condition, \Cref{tab:trajectory-source-ablation} reports the source's success rate on the 500 collection tasks and the test pass@1 of the resulting student under $h$.

\noindent\textbf{Obs.\ding{187} Effective supervision comes from harness-guided review, not from stronger demonstrations alone.}
Directly fine-tuning on GPT-5.6 Sol trajectories leaves the student at the 12\% base result, even though that source succeeds on 52.0\% of the collection tasks.
Likewise, trajectories collected by the base student under $h^\star$ yield 12\% after SFT.
This comparison isolates the importance of starting from the student's trajectory and translating harness guidance into corrections compatible with its current state and target action space.

\noindent\textbf{Obs.\ding{188} Procedural guidance provides better distillation supervision than answer access or generic review.}
Review with an empty $\mathcal K$ reaches only 11\%.
Providing oracle answers raises collection success to 98.6\%, yet the distilled model reaches only 15\%.
By comparison, $\mathcal K$ contains no task answers and reaches a lower collection success of 59.4\%, yet the student distilled from its trajectories reaches 30\% pass@1.
Collection success therefore does not predict distillation value. Oracle access encourages non-generalizable shortcut corrections.
By contrast, $\mathcal K$ instills reusable procedural behavior (e.g., reasoning from reaction templates, proposing candidates, and systematically validating reactant sets) that the student can execute independently under $h$.

\noindent\textbf{Obs.\ding{189} Executing $h^\star$ during collection does not make its behavior transferable.}
Mounting $h^\star$ improves GPT-5.6 Sol's collection success from 52.0\% to 62.0\%, but the resulting distilled student falls from 12\% to 3\%.
The student-generated counterpart also succeeds on 39.4\% of collection tasks but returns to 12\% after SFT.
The failure is consistent with action-space mismatch. The model distilled from teacher trajectories under $h^\star$ extensively attempts unavailable harness-tool calls, and many trials thus exhaust the turn limit.
\our{} avoids this mismatch by expressing each correction through $h$.

\begin{wraptable}{r}{0.5\textwidth}
    \centering
    \caption{Comparing alternative distillation signals on USPTO.}
    \label{tab:trajectory-source-ablation}
    \footnotesize
    \setlength{\tabcolsep}{2pt}
    \renewcommand{\arraystretch}{1.1}
    \begin{tabular}{lcc}
    \Xhline{1.1pt}
    \rowcolor{hzheader}
    \textbf{Trajectory source}
    & \makecell{\textbf{Collection}\\\textbf{success}}
    & \makecell{\textbf{Test pass@1}\\\textbf{under $h$}} \\
    \Xhline{1.1pt}
    Base (untrained) & --- & 12.0 \\
    \rowcolor{hzgroup}
    \multicolumn{3}{c}{\textit{Direct distillation}} \\
    Teacher rollout & 52.0 & 12.0\nochange{0.0} \\
    Teacher under $h^\star$ & 62.0 & 3.0\loss{9.0} \\
    Student under $h^\star$ & 39.4 & 12.0\nochange{0.0} \\
    \rowcolor{hzgroup}
    \multicolumn{3}{c}{\textit{Agent-as-harness}} \\
    Review with empty $\mathcal K$ & 44.2 & 11.0\loss{1.0} \\
    Review with oracle answer & 98.6 & \snd{15.0}\gain{3.0} \\
    \textbf{\our{} ($\mathcal K$)} & 59.4 & \textbf{30.0}\gain{18.0} \\
    \Xhline{1.1pt}
    \end{tabular}
\end{wraptable}

\paragraph{We discuss that two factors make $\mathcal K$-guided step-level review more effective.}
First, smaller models struggle to internalize guiding steps from a stronger model when those steps rely on unconstrained reasoning beyond the student's capability; by contrast, $\mathcal K$ translates the evolved harness into standardized, tool-grounded procedures that are much easier for a student to learn.
Second, standalone teacher rollouts are off-policy and rarely enter the intermediate failure states that the student encounters at test time, leaving the student unable to recover from its own errors.
In \Cref{app:data-collection} (\Cref{tab:matched-controls}), we further extend these ablations to SpreadsheetBench and AppWorld, showing that \our{} substantially outperforms student rejection-sampling SFT with a $2\times$ rwollout budget, distillation from standalone teacher rollouts, and distillation from rollouts reviewed with an empty $\mathcal K$.

\subsection{Behavioral internalization}\label{sec:exp-behavior}

To evaluate harness distillation at a finer granularity, we measure how well the distilled model internalizes the behavioral patterns encoded in $h^\star$.
We first translate patterns of enabled tools, middleware, memory, and skills in $h^\star$ into trajectory detectors.
For each detector, we compare the base model's trajectories under $h^\star$ and $h$ on the same test task.
We select tasks where the pattern appears only under $h^\star$ and keep a pattern only when at least 10 tasks meet this criterion.
This yields 18 harness-exclusive patterns on SpreadsheetBench, 6 on USPTO, and 4 on AppWorld; \Cref{app:behavior-patterns} details the mining procedure and explains every pattern.
For each pattern, its \emph{recovery rate} is the fraction of the selected tasks on which \our{} exhibits the same behavior.

\begin{table}[t]
\centering
\scriptsize
\renewcommand{\arraystretch}{1.08}
\caption{Recovery of 28 harness-exclusive behaviors across three domains.}
\label{tab:behavior-internalization}
\begin{minipage}[t]{0.46\linewidth}
\centering
\begin{tabularx}{\linewidth}{l>{\raggedright\arraybackslash}Xc}
\Xhline{1.1pt}
\rowcolor{hzheader}
\textbf{Source} & \textbf{Pattern} & \textbf{Recovery} \\
\Xhline{1.1pt}
\rowcolor{hzgroup}
\multicolumn{3}{c}{\textbf{SpreadsheetBench} (18 patterns)} \\
\multirow{4}{*}{Memory}
& Edit existing workbook       & 31/34 (91\%) \\
& Assert workbook structure    & 20/27 (74\%) \\
& Avoid fragile coordinates    & 19/19 (100\%) \\
& Save requested workbook      & 14/14 (100\%) \\
\addlinespace[1pt]
\multirow{8}{*}{Skill}
& Inspect before editing       & 32/32 (100\%) \\
& Inspect target/examples      & 14/18 (78\%) \\
& Use \texttt{solution.py}     & 53/76 (70\%) \\
& Execute final script         & 52/75 (69\%) \\
& Read cached formulas         & 28/43 (65\%) \\
& Normalize matching keys      & 6/10 (60\%) \\
& Use closed target ranges     & 30/32 (94\%) \\
& Reload saved workbook        & 17/17 (100\%) \\
\addlinespace[1pt]
\multirow{4}{*}{Middleware}
& Protect pre-filled cells     & 24/26 (92\%) \\
& Write computed literals      & 19/21 (90\%) \\
& Populate answer range        & 28/30 (93\%) \\
& Scan range after save        & 20/24 (83\%) \\
\Xhline{1.1pt}
\end{tabularx}
\end{minipage}\hfill
\begin{minipage}[t]{0.50\linewidth}
\centering
\begin{tabularx}{\linewidth}{l>{\raggedright\arraybackslash}Xc}
\Xhline{1.1pt}
\rowcolor{hzheader}
\textbf{Source} & \textbf{Pattern} & \textbf{Recovery} \\
\Xhline{1.1pt}
\rowcolor{hzgroup}
\multicolumn{3}{c}{\textbf{SpreadsheetBench} (cont.)} \\
\multirow{2}{*}{Middleware}
& Check output existence       & 16/16 (100\%) \\
& Stop after delivery          & 21/28 (75\%) \\
\addlinespace[2pt]
\rowcolor{hzgroup}
\multicolumn{3}{c}{\textbf{USPTO} (6 patterns)} \\
\multirow{2}{*}{Skill}
& Parse product with RDKit first & 23/23 (100\%) \\
& Enumerate candidate sets       & 59/88 (67\%) \\
\addlinespace[1pt]
\multirow{3}{*}{Tool}
& Validate with RDKit             & 60/60 (100\%) \\
& Canonicalize SMILES             & 73/73 (100\%) \\
& Check heavy-atom coverage       & 40/54 (74\%) \\
\addlinespace[1pt]
Middleware
& Validate the final answer       & 45/47 (96\%) \\
\addlinespace[2pt]
\rowcolor{hzgroup}
\multicolumn{3}{c}{\textbf{AppWorld} (4 patterns)} \\
\multirow{2}{*}{Skill}
& Retrieve complete pages         & 21/53 (40\%) \\
& Read back mutations             & 6/12 (50\%) \\
\addlinespace[1pt]
\multirow{2}{*}{Middleware}
& Avoid repeated failed calls     & 22/31 (71\%) \\
& Choose action vs.\ answer       & 24/33 (73\%) \\
\Xhline{1.1pt}
\end{tabularx}
\end{minipage}
\end{table}

\noindent\textbf{Obs.\ding{190} \our{} internalizes behaviors contributed by $h^\star$.}
Averaged over the 28 patterns in \Cref{tab:behavior-internalization}, \our{} recovers 82.3\% of the harness-exclusive behaviors.
Recovery spans memory, skill, tool, and middleware sources, covering both model-visible guidance and executable components.
Each pattern is scored only on the tasks selected for it, where the base model exhibits the behavior under $h^\star$ but never under $h$, so the base model scores 0\% on every pattern by construction.
These recovery rates therefore measure behavior that distillation adds, providing direct evidence that \our{} internalizes behavior induced by $h^\star$.

\section{Conclusion and discussion}\label{sec:conclusion}

\our{} turns an evolved student-side harness $h^\star$ into training supervision for a model operating under a fixed target harness $h$. Guided by the adapted reference harness $\mathcal K$, a harnessing agent rewrites the student's proposals before execution, producing training trajectories compatible with $h$.
Across three domains, agent-as-harness outperforms code-as-harness on frontier models (81.1\% vs.\ 78.1\% on average), and ablations identify harness-guided review as the most effective of the tested supervision sources.
After SFT, the student under $h$ alone raises macro-average task success from 23.3\% to 44.3\%, surpassing the 41.7\% achieved with $h^\star$. 
In addition, it reduces per-task inference token usage by 27\%--86\% compared with the base model under $h^\star$ and recovers $h^\star$-specific behaviors at an average rate of 82.3\% across 28 patterns.

\paragraph{Limitations.}
The method depends on a capable harnessing model. With weaker models, review can become harmful and agent-as-harness loses its advantage over code-as-harness (\Cref{app:aah-analysis}). Reviewing every proposal also increases collection cost; each step requires an additional model call that processes the trajectory and $\mathcal K$, raising mean USPTO latency to $2.4\times$ during collection, an overhead absent after distillation. SFT may also fail to fully internalize deep domain knowledge encoded by $h^\star$.
In addition, some harness mechanisms, e.g., context management, are not fully expressible as student responses out of the box.
\our{} therefore does not remove the need for a harness, but narrows what that harness must provide, and we expect a minimal one to suffice as harness distillation improves.

\paragraph{Future work.}
A harnessing agent must anticipate how executing the student's proposal would change the environment and whether an intervention would yield a better trajectory, so effective harnessing depends on an accurate model of agent--environment dynamics; training stronger agent world models~\citep{zuo2026qwenagentworld} and specializing them for harnessing decisions is a promising model-level direction.
At the framework level, reviewing every proposal and restricting intervention to full-response replacement incurs unnecessary calls and provides only coarse-grained control. Future work should establish proxy signals to invoke review selectively and leverage finer-grained harnessing such as token insertion and latent-space steering.
The training signal can also be improved. Each replacement pairs a rejected and a preferred response at the same state, so preference learning could exploit comparison signals that response-level SFT discards.

Overall, \our{} paves the way for recursive self-improvement based on agent harnesses: distilling task-specific harnesses into shared weights lets behaviors and knowledge accumulate in parameters, so better models build better harnesses and each harness returns its gains to the weights.

\section*{AI Use Statement}

We used AI assistants in two roles. First, to check grammar and to polish text the authors had written. Second, for routine coding assistance during implementation. The method, the experimental design, and every implementation decision affecting the reported results were made by the authors, who verified all AI-assisted output and take full responsibility for this paper.

\bibliography{main}

\clearpage
\appendix

\vbox{%
  \hsize\textwidth
  \linewidth\hsize
  \vskip 0.1in
  \hrule height 4pt
  \vskip 0.25in
  \vskip -\parskip
  \centering
  {\LARGE\bf \titleour:\\Harness Distillation via Agent-as-Harness\\[0.15cm](Appendix)}
  \vskip 0.29in
  \vskip -\parskip
  \hrule height 1pt
  \vskip 0.09in
}

\startcontents[appendix]
\vspace{0.5cm}
\printcontents[appendix]{}{1}{\setcounter{tocdepth}{2}}
\vspace{1cm}

\clearpage

\section{Skill-guided harness evolution}\label{app:harness-evolution}

\subsection{Evolution methods}

We evolve one shared student-side harness for each benchmark. The process, also known as meta-harness, begins with rollouts under $h$ on the training split. In each round, a swarm of analysis agents examines the tasks that failed in the preceding round and proposes reusable changes to tools, middleware, skills, or memory. A main evolution agent consolidates these proposals into the shared harness and evaluates the same model with that harness attached. Tasks that pass leave the evolution process, while the remaining failures form the next round. We run this process for three rounds.

The evolution objective is performance under the student-side harness itself. Evaluation therefore uses the target student agent with the evolved harness attached and no harnessing-agent intervention. Proposed components must address failure patterns shared across tasks. \Cref{lst:harness-evolution-skill} shows the skill followed by the evolution agent.

\begin{lstlisting}[style=hzlisting]
---
name: student-harness-evolve
description: Generic workflow for evolving a shared student harness -- for a benchmark's Harbor-format train task set, take bare miniswe rollouts as the baseline, handle only the tasks the previous round got wrong, use a swarm (default 10 coder subagents) to analyze rollout trajectories in parallel, have the main agent write the shared student harness (tools / middleware / skills / memory) under `harness_bank/<domain>/`, and iterate for 3 rounds with "miniswe eval rollouts loading that harness" as the evaluator. The evolved artifact can later be rewritten into a teacher-side harness.
---

# Shared Student Harness Evolution Guide

## What this workflow does

For a benchmark's train task set, evolve **one** deliverable: a **student-side harness shared by all tasks**, kept in a per-benchmark bank (`harness_bank/<domain>/`):

- `tools/` -- prebuilt tools (real StructuredTools loaded into miniswe alongside `execute`);
- `middlewares/` -- student-side langchain AgentMiddleware;
- `skills/<name>/SKILL.md` + `memory.md` -- skill and memory material, uploaded into the sandbox and exposed to the student via deepagents' SkillsMiddleware / MemoryMiddleware.

**The evaluator** is a bare miniswe rollout with the harness loaded (`ahd.harness:AHDMinisweAgent`, teacher passthrough, i.e. no teacher intervention; the bank is mounted via `student_harness_dir`). It measures "does this student harness help the model itself". A campaign runs **3 rounds** on the rhythm: "select the tasks the previous round got wrong -> parallel swarm analysis -> main agent writes the bank -> eval rollout on exactly those failed tasks -> attribution". Tasks already solved count as "the harness is good enough for them" and are not rolled out again in later rounds.

## Campaign parameters (set at each instantiation)

| Parameter | Meaning |
|---|---|
| `<DOMAIN>` | Domain name (determines the bank path `harness_bank/<domain>/`; for the student side, prefer a `<benchmark>_student` suffix to distinguish it from teacher-side banks) |
| `<TASKS>` | Harbor task path + task-list file (e.g. `-p data/spreadsheetbench-verified`, task list `experiments/spreadsheetbench_train300.txt`) |
| `<MODEL>` | Evaluation model (e.g. `openrouter:qwen/qwen3.5-9b`, reasoning explicitly enabled) |
| `<BASELINE_JOBS>` | Job directory of the baseline (bare, no harness) rollout; the input for round-1 failure analysis. If it does not exist, run the baseline first |
| `<N_AGENTS>` | Swarm concurrency, at most 10 subagents at a time; all of the round's failed tasks are distributed among these subagents, with no per-subagent task cap |

## Bank structure and integration

```
harness_bank/<domain>/
|-- registry.py        # single entry point; API contract below
|-- manifest.json      # enablement manifest: tools/middlewares/skills lists + memory switch; toggle components only here
|-- memory.md          # accumulated checklist clauses (English, no source attribution, grouped by topic)
|-- tools/             # *.py, each exposing make_tool(backend) -> StructuredTool
|-- middlewares/       # *.py, each exposing make_middleware() -> AgentMiddleware
|-- skills/            # <name>/SKILL.md, deepagents skill format (progressive disclosure)
`-- README.md          # bank description and provenance
```

- **Registry API contract** (consumed by the loader under exactly these names): `TOOL_FACTORIES` / `MIDDLEWARE_FACTORIES` (dict, name -> `"module:function"` import string, which the loader resolves into a factory). `catalog_text()` (component listing rendering) is only for logs and audits; the loader does not consume it. Component docstrings are the catalog source.
- **Loading semantics** (the contract of `student_harness_dir`, implemented in `src/ahd/student_harness.py` / `src/ahd/student.py`):
  1. tools are appended to miniswe's tools list (after `execute`);
  2. the `skills/` directory and memory.md are uploaded into the sandbox (`/opt/ahd/harness/skills/`, `/opt/ahd/harness/memory.md`): skills are progressively disclosed through deepagents' `SkillsMiddleware` (only the index goes into the system prompt; the student reads full text via `execute cat`), and memory.md is injected into the system prompt through `MemoryMiddleware`; SkillsMiddleware and MemoryMiddleware run **before** the review middleware so the teacher sees the same context as the student;
  3. the bank's middleware is appended after the review middleware, followed by `StudentRequestCaptureMiddleware` (records the final model request, for audit);
  4. **bank bytes are hashed into `bank_sha256`** -- the rollout log writes `student_harness_bank.txt` recording the mounted bank's hash, for provenance only, with no assertions;
  5. when `student_harness_dir` is not passed, behavior is exactly the status quo (bare compatible).
- **The student core is not evolvable**: the `execute` tool, the student.md base prompt, the review middleware, the turn limit, and the tool_call adapter are fixed layers; evolution happens only in bank content.

## Per-round workflow

### Phase 0: determine this round's failed-task set

- Round 1: select tasks whose reward did not pass from the `<BASELINE_JOBS>` results (if the baseline does not exist, first run bare rollouts over the train task list with `<MODEL>`).
- Rounds 2/3: select tasks whose reward did not pass from the previous round's harnessed eval rollout.
- This round's swarm, bank edits, and eval rollout cover only these failed tasks. Tasks already solved leave the campaign.
- If the failed-task set is empty, end the campaign early; completed rounds still count as valid results.

### Phase 1: swarm (at most 10 coder subagents at a time; analyze and propose changes only)

Launch with AgentSwarm, item = this round's failed-task group. Each subagent does two things for its assigned tasks:

1. **Failure analysis -> `experiments/<domain>_evolution/round<N>/notes/<TASK>.md`**
   - Round 1: read the task's bare trajectory in `<BASELINE_JOBS>` (`<job>/<TASK>__*/agent/trajectory.json`, `llm_calls.jsonl`, `verifier/`), and understand how the agent worked and the failure class (insufficient exploration / mechanism misused / miscalculation / deliverable missing or misplaced / missing verification discipline).
   - Rounds 2/3: read the previous round's harnessed trajectory and attribute per the Attribution checklist section.
2. **Propose bank changes**: per Evolution discipline, decide whether the failure should be addressed by generalizing an existing component, adding a new component, or editing prompt/memory material; write the proposal and its rationale in the notes (it must argue "this helps a class of tasks"); do not write to `harness_bank/` directly.

Subagent forbidden zone: do not modify `harness_bank/`, `src/`, or `data/`; do not run rollouts; do not perform git operations.

**WARNING -- component hard rule:** middleware hooks run on the harbor event loop; sandbox probing may only use `await backend.aexecute(...)`; never call the synchronous `backend.execute(...)` inside an async hook -- internally it is `run_coroutine_threadsafe(...).result()`, which deadlocks the entire event loop when called on the loop thread (0% CPU, all trials frozen, even trial-level timeouts cannot fire). Tool function bodies run on worker threads and are not subject to this restriction; message-level checks in middleware are always the safest choice.

### Phase 2: main-agent write-up and acceptance

1. **Consolidated writing**: read all of the round's notes, resolve duplicate or conflicting proposals, and have the main agent uniformly modify the bank (tools / middleware / skills / memory.md / manifest.json).
2. **Reconciliation**: grep every reference name in manifest.json against the registered names in the registry; stale references must be zeroed out (otherwise loading fails outright).
3. **Smoke test**: import smoke (`from harness_bank.<domain>.registry import ...` + render the catalog); `load_student_harness(bank_dir)` resolves successfully (all manifest references hit, `skills/<name>/SKILL.md` files all present); `pytest tests/ -q` shows no regressions.
4. **Eval rollout**: submit a rollout of bare miniswe with the bank loaded, using this round's failed-task IDs as the exact include set (teacher passthrough, same model as the baseline, job name carries a date tag such as `YYYYMMDD-<domain>-evol-r<N>`). Before the campaign starts, review with the user once per AGENTS.md: the full train task list, the failure-selection criterion, the number of rounds, the fixed model and hyperparameters, and a directly executable command template; after user confirmation, later rounds proceed automatically under the confirmed rules without asking again each round. Re-review whenever the model, dataset, selection criterion, or hyperparameters change.

Reading results: the reward distribution in `runs/<job>/result.json`; per-trial details in `<job>/<trial>/{agent,verifier}`. Immediately after the rollout finishes, generate the next round's failed-task list and record the tasks that passed this round and left the campaign.

## Evolution discipline

1. **The lib is the only layer.** Shared mechanism -> bank; specific to one task -> do not write it, accept the residual failure. The one-sentence test: is this failure "shared by a class of tasks" or "specific to this one task" -- shared -> bank; specific -> give up on that task, and never write task-specific content (concrete cell addresses, concrete answers, steps unique to one task, concrete file names) into any shared component, skill, or memory. This is both overfitting prevention and the foundation of the claim that "the harness is generic".
2. **Generalize before adding.** If attribution points to a mechanism the bank already has but that does not quite fit -> generalize and improve the existing component (near-duplicate variants are forbidden); if it truly does not exist -> write a new component (first ask yourself "would this be useful on other tasks"; only if generic does it enter bank + manifest + registry).
3. **Prompt material is also a shared asset.** For behavioral failures (finishing without verification, acting before reading the task, writing the deliverable to the wrong place), prefer editing the corresponding `skills/<name>/SKILL.md` or appending generic clauses to memory.md (English, no source attribution, merged into the matching topic section); do not write task-specific exhortations.
4. **Anti-bloat discipline.** Near-duplicate proposals are deduplicated by the main agent during consolidation; component survival is adjudicated by evaluator performance -- components with no evidence of benefit for two consecutive rounds are moved out of the manifest (files kept for traceability).
5. **Traces are evidence.** Every change must point back to a trajectory attribution in the notes; components that "feel like they should help" are not allowed.

## Attribution checklist (round >=2, go through in order, record in notes)

1. **Did the harness load?** -> In the trial log, confirm the bank_sha256 in `student_harness_bank.txt` matches the current bank and that the skills index appears in the system prompt; if not mounted: manifest/registry reconciliation or a loader problem.
2. **Did the student use the evolved tools?** -> Search `llm_calls.jsonl` for the evolved tools' names; if unused: the tool description is not discoverable or the skills index in the system prompt gives no guidance -- fix descriptions / material rather than adding new tools.
3. **Did the middleware fire?** -> If it fired but did not help: the clauses are not actionable enough -- rewrite abstract principles into verbatim rules; if it did not fire: the hook condition does not match the actual trajectory shape.
4. **Is the student dying on mechanics?** (hangs / malformed tool calls / validation loops / repeatedly retrying a failing tool) -> mechanics problem, go back to Evolution discipline and fix the bank.
5. **Dead-trial classification**: infra (does not count) vs turn wall (non-convergence, attribution 2/3/4) vs verifier failure (real failure -- read the verifier details and distinguish "wrong value / wrong location / missing deliverable / wrong format").

## Rounds and acceptance

- Run 3 rounds in total; end early if the failed-task set empties. Record per round: number of input failed tasks, number of tasks that passed this round and left, number of remaining failed tasks, mean reward (vs the bare baseline on the same task set), component additions/changes, recurring failure patterns, and the composition of abnormal trials (infra deaths vs real failures).
- Low scores in round 1 are expected (the v1 artifact comes from static analysis alone, with no attribution iterations); watch the trend, not the absolute value.
- Final delivery after all rounds: the reward curve, the bank's final-state inventory (manifest + component catalog), a map of failure patterns, open issues (including the list of abandoned task-specific residual failures), and suggested material for rewriting into a teacher-side harness.
\end{lstlisting}
\captionof{skilllisting}{Harness evolution skill.}\label{lst:harness-evolution-skill}

\subsection{Evolved components}\label{app:hstar-inventory}

\Cref{tab:hstar-inventory} lists every harness component that survives in the final $h^\star$ of each domain. SpreadsheetBench and USPTO keep their accumulated failure notes in an enabled memory component, whereas AppWorld disables the memory slot and injects the same material through the \texttt{bootstrap\_instruction} middleware, which is why the table lists no memory row for it.

\begin{table}[!p]
\centering
\hztablesetup
\caption{Component inventory of the evolved student-side harness $h^\star$ in each domain.}
\label{tab:hstar-inventory}
\begin{tabular}{lp{0.80\linewidth}}
\hztablerule
\rowcolor{hzheader}
\multicolumn{1}{c}{\textbf{Type}} & \multicolumn{1}{c}{\textbf{Component and function}} \\
\hztablerule
\rowcolor{hzgroup}
\multicolumn{2}{c}{\textbf{SpreadsheetBench}} \\
Memory & \texttt{memory.md}. Seven hard rules: read the skill before exploring, deliver the input workbook edited in place, treat pre-filled cells and worked examples as the specification, write computed literals, address cells by Excel coordinates, verify by assertion against independent recomputation, and remember that the deliverable is always the workbook. \\
Skill & \texttt{spreadsheet-manipulation}. Workflow and pitfall catalog for \texttt{.xlsx} editing: explore before editing, write literal values instead of formulas, read formula cells with \texttt{data\_only}, normalize lookup keys, keep the target range clean. \\
Middleware & \texttt{empty\_turn\_guard}. Reject a finish whose assistant message carries neither a tool call nor visible content. \\
Middleware & \texttt{output\_guard}. Reject a finish when no non-empty workbook exists in the output directory. \\
Middleware & \texttt{prefilled\_guard}. Diff the input and output workbooks inside the answer range and reject a finish that overwrote a pre-filled literal cell. \\
Middleware & \texttt{answer\_range\_guard}. Mechanical health check of the deliverable range: all-empty range, formula strings, numbers stored as text, error literals, degenerate constant fills. \\
Middleware & \texttt{turn\_budget}. Advise on remaining budget at turns 25 and 35, and interrupt a loop that repeats the same exception. \\
\addlinespace[3pt]
\rowcolor{hzgroup}
\multicolumn{2}{c}{\textbf{AppWorld}} \\
Skill & \texttt{appworld-workflow}. Public-API helpers: strict parameters, per-app authentication, complete pagination, verified mutations, explicit action and answer completion. \\
Middleware & \texttt{bootstrap\_instruction}. Load the workflow helper and keep one concise guide visible in the system prompt. \\
Middleware & \texttt{failure\_recovery\_instruction}. After a failed execution, supply concrete recovery steps for that specific failure. \\
Middleware & \texttt{context\_guard}. Bound tool observations before they enter message history or the next model request. \\
Middleware & \texttt{submit\_gate}. Nudge the student when an observed supervisor status reports the task unfinished. \\
\addlinespace[3pt]
\rowcolor{hzgroup}
\multicolumn{2}{c}{\textbf{USPTO}} \\
Memory & \texttt{memory.md}. Checklist clauses in three groups. (1) \textit{Answer discipline.} The grader matches strings exactly, so ship RDKit-canonical SMILES without atom maps, write a best-guess answer file first and refine it in place, and validate after the file's last write. (2) \textit{Mechanical verification.} Every fragment parses, reactant heavy atoms account for every product heavy atom, and each fragment overlaps the product almost completely. (3) \textit{Chemical priors.} The recorded set contains every species contributing heavy atoms, the stated reaction type is a hard filter over candidate transforms, and ties break toward the smallest atom edit. \\
Skill & \texttt{retrosynthesis-candidates}. Enumerate candidate disconnections for the named reaction, filter them mechanically, and commit to one; reactant-set conventions and representation pitfalls that score zero like a wrong molecule. \\
Tool & \texttt{propose\_retrosynthesis}. Enumerate candidate reactant sets with a curated library of named one-step transforms. \\
Tool & \texttt{smiles\_check}. RDKit-backed per-candidate scoring: validity, atom budget, maximum-common-substructure coverage, changed reaction sites. \\
Tool & \texttt{verify\_answer}. Output-contract battery over the answer file, plus an RDKit-canonical rewrite of the submitted SMILES. \\
Middleware & \texttt{answer\_guard}. Reject a finish when the answer file is missing, empty, or grossly malformed. \\
Middleware & \texttt{completeness\_guard}. Compare product and reactant heavy-atom counts with RDKit and reject a finish whose reactant set is missing a co-reactant. \\
\hztablerule
\end{tabular}
\end{table}

\section{Harness adaptation}\label{app:harness-adaptation}

The adaptation from $h^\star$ to $\mathcal K$ preserves what each component does while changing how and where it acts. \Cref{tab:translation} summarizes the component-level mapping. Beyond individual components, the adaptation can also produce a domain prompt for the harnessing agent that states the domain's review policy.

\begin{table}[!ht]
\centering
\hztablesetup
\caption{Component-level adaptation from the evolved student-side harness to the reference harness.}
\label{tab:translation}
\begin{tabular}{p{0.45\linewidth}p{0.45\linewidth}}
\hztablerule
\rowcolor{hzheader}
\multicolumn{1}{c}{\textbf{Student-side harness $h^\star$}} & \multicolumn{1}{c}{\textbf{Private reference harness $\mathcal K$}} \\
\hztablerule
\textbf{Tool.} Expose a utility directly in the student's action space.
&
\textbf{Action recipe.} Specify how to construct an equivalent action that the student can execute through $h$. \\
\addlinespace[2pt]
\textbf{Middleware.} Inspect the student's state or execution and modify or block the loop when a condition is met.
&
\textbf{Review middleware.} Detect the corresponding condition from the proposal and visible trajectory, then privately alert the harnessing agent. \\
\addlinespace[2pt]
\textbf{Skill.} Give the student a workflow for inspecting, editing, and verifying its work.
&
\textbf{Review guidance.} Use the workflow to diagnose the current step and construct a student-native replacement when needed. \\
\addlinespace[2pt]
\textbf{Memory.} Record recurring failures and the practices that avoid them.
&
\textbf{Failure patterns.} Turn each failure into a recognizable review condition and a targeted intervention. \\
\hztablerule
\end{tabular}
\end{table}

\subsection{Reference harness inventory}\label{app:k-inventory}

$\mathcal K$ is mounted as a read-only \texttt{/components} directory in the harnessing agent's own workspace, which is separate from the student sandbox. Its entry point is an \texttt{index.json} that lists every readable component with an identifier, a kind, a relative path, and a one-line summary. The harnessing agent reads the index before its first decision and then opens individual component files as they become relevant. \Cref{tab:k-inventory} gives the full component inventory for each domain.

\begin{table}[!ht]
\centering
\hztablesetup
\setlength{\tabcolsep}{3pt}
\renewcommand{\arraystretch}{1.08}
\caption{Component inventory of the private reference harness $\mathcal K$ in each domain.}
\label{tab:k-inventory}
\begin{tabular}{lp{0.78\linewidth}}
\hztablerule
\rowcolor{hzheader}
\multicolumn{1}{c}{\textbf{Type}} & \multicolumn{1}{c}{\textbf{Component and content}} \\
\hztablerule
\rowcolor{hzgroup}
\multicolumn{2}{c}{\textbf{SpreadsheetBench} (8 components)} \\
Failure patterns & \texttt{accumulated-failures}. Recurring student mistakes on spreadsheet tasks and how to correct them, such as editing in place and preserving pre-filled cells. \\
Review guidance & \texttt{spreadsheet-manipulation}. The correct \texttt{.xlsx} editing workflow and its pitfalls; used to diagnose what the current proposal got wrong. \\
Action recipe & \texttt{prefilled-diff}. How to build a student-native command that checks whether any pre-filled example cell was overwritten. \\
Action recipe & \texttt{answer-range-check}. How to build a student-native command that sanity-checks the answer range before finishing. \\
Review middleware & \texttt{stall}. Catches proposals that declare intent without acting, and calls for a concrete action. \\
Review middleware & \texttt{error-loop}. Catches the student repeating the same failing command, and calls for consulting the documentation instead of retrying. \\
Review middleware & \texttt{turn-budget}. Reminds the harnessing agent late in the trial to steer the student toward finishing. \\
Review middleware & \texttt{finish-guard}. Catches a finish submitted without the verification steps. \\
\addlinespace[3pt]
\rowcolor{hzgroup}
\multicolumn{2}{c}{\textbf{AppWorld} (6 components)} \\
Failure patterns & \texttt{appworld-review}. Recurring review failures on AppWorld tasks, such as inexact entities or metrics and submission without evidence. \\
Review guidance & \texttt{api-workflow}. How to work with the public APIs correctly: contracts, credentials, and pagination. \\
Review guidance & \texttt{actions-files}. How to mutate app state safely: act on original identifiers, preserve files, and verify before deleting or completing. \\
Action recipe & \texttt{public-patterns}. How to express complete pagination and app-file read-back as ordinary student \texttt{execute} commands. \\
Review middleware & \texttt{appworld-review-v1}. A battery of checks on each proposal, covering malformed responses, failed executions, premature submission, and unsafe writes. \\
Domain prompt & \texttt{teacher-prompt}. Domain review rules appended to the harnessing agent's system prompt: the student's exact action space, and how to spend interventions on substantive fixes. \\
\addlinespace[3pt]
\rowcolor{hzgroup}
\multicolumn{2}{c}{\textbf{USPTO} (13 components)} \\
Failure patterns & \texttt{accumulated-failures}. Recurring student mistakes on retrosynthesis tasks and how to correct them, such as ranking candidates on graph evidence without chemical priors. \\
Review guidance & \texttt{retrosynthesis-candidates}. How to compare candidate reactant sets on reaction evidence instead of superficial heuristics. \\
Action recipe & \texttt{propose-retrosynthesis}. Reference implementation for enumerating candidate reactant sets. \\
Action recipe & \texttt{smiles-check}. Reference implementation for scoring each candidate mechanically. \\
Action recipe & \texttt{verify-answer}. Reference implementation for checking the output contract and rewriting the answer in RDKit-canonical form. \\
Review middleware & \texttt{candidate-format}. Catches structurally invalid proposals, which a \textsc{pass} would preserve. \\
Review middleware & \texttt{command-timeout}. Catches commands that could block indefinitely, and calls for a shell timeout. \\
Review middleware & \texttt{enumeration}. Catches a hand-written answer file, the signature of skipping mechanical candidate enumeration. \\
Review middleware & \texttt{canonical-form}. Catches answer files not produced by the canonicalizing verification script. \\
Review middleware & \texttt{stall}. The same intent-without-action check as in SpreadsheetBench. \\
Review middleware & \texttt{answer-file}. Catches a finish without evidence that the answer file was written and verified. \\
Review middleware & \texttt{completeness}. Catches a finish whose reactant set may not account for every heavy atom in the product. \\
Domain prompt & \texttt{teacher-prompt}. Domain review rules appended to the harnessing agent's system prompt: the stepwise derivation every trajectory must follow, and when to intervene. \\
\hztablerule
\end{tabular}
\end{table}

Review middleware is not reachable as files. These components run automatically on every proposal and, when a condition matches, append a short block of triggered guidance to the update the harnessing agent receives. The domain prompt is appended directly to the harnessing agent's system prompt.

The adaptation is not one-to-one. For example, a student-side guard that probes the sandbox becomes a review-time condition on the visible trajectory, plus an action recipe whenever the correction itself must run in the environment. One guard can also split into several checks. USPTO's \texttt{answer\_guard} is translated into separate conditions for a missing answer file, a hand-written file that bypasses canonicalization, and a commitment made without candidate enumeration.

\subsection{Worked example: preserving pre-filled cells}

Many SpreadsheetBench tasks place a few already-filled example cells inside the requested answer range. Those cells are the specification. The student should infer the filling rule from them, match their format, and write only into the remaining blanks. Overwriting an example is a common failure. The evolved harness therefore inspects the sandbox at finish time, diffs the input and output workbooks inside the answer range, and rejects the finish if any pre-filled literal has changed. \Cref{lst:student-prefilled-guard} shows this control flow.

\begin{lstlisting}[style=hzlisting,language=Python]
async def on_proposed_finish(state):
    input_path, answer_range, output_path = parse_task(state)
    result = await run_in_student_sandbox(
        PREFILLED_DIFF, input_path, output_path, answer_range
    )
    if result.violations:
        reject_finish(
            cells=result.violations,
            instruction="Restore the examples and re-derive the rule."
        )
\end{lstlisting}
\captionof{codelisting}{Abridged student-side middleware in $h^\star$.}\label{lst:student-prefilled-guard}

The harnessing agent cannot run this check privately because it has no separate interface to the student environment, and it should not have one. Otherwise the behavior patterns behind this middleware cannot be internalized. The adapted review middleware instead looks for evidence of the comparison in the student-visible trajectory. If the student tries to finish without that evidence, the middleware adds a private instruction asking the harnessing agent to replace the finish with a verification action:

\begin{lstlisting}[style=hzlisting,language=Python]
def review(candidate, visible_trajectory):
    if not candidate.is_finish():
        return None

    evidence = scan_student_actions(visible_trajectory)
    if not evidence.has_prefilled_comparison:
        return PrivateInstruction(
            decision="REPLACE",
            action="Run the pre-filled-cell comparison before finishing."
        )
\end{lstlisting}
\captionof{codelisting}{Abridged review middleware in $\mathcal K$.}\label{lst:teacher-finish-guard}

The corresponding tool component in $\mathcal K$ takes the form of an action recipe. It contains the logic needed to generate a utility script inside the accepted student response. A shortened version is shown in \Cref{lst:prefilled-diff}. The response creates and invokes the script through $h$; its output then becomes visible to the student on the next turn.

\begin{lstlisting}[style=hzlisting,language=bash]
cat > .agent-tools/prefilled_diff.py <<'PY'
import sys
import openpyxl
from openpyxl.utils import range_boundaries

input_path, output_path, cell_range = sys.argv[1:4]
wb_in = openpyxl.load_workbook(input_path)
wb_out = openpyxl.load_workbook(output_path)
ws_in, ws_out = wb_in.active, wb_out.active
min_c, min_r, max_c, max_r = range_boundaries(cell_range)

changed = []
for row in range(min_r, max_r + 1):
    for col in range(min_c, max_c + 1):
        before = ws_in.cell(row=row, column=col)
        after = ws_out.cell(row=row, column=col)
        if before.value is not None and before.data_type != "f":
            if after.value != before.value:
                changed.append(before.coordinate)

print("PASS" if not changed else f"FAIL changed cells: {changed}")
raise SystemExit(bool(changed))
PY
python3 .agent-tools/prefilled_diff.py INPUT.xlsx OUTPUT.xlsx B3:B40
\end{lstlisting}
\captionof{codelisting}{Student-native verification action generated from the action recipe in $\mathcal K$.}\label{lst:prefilled-diff}

\subsection{Adapting skills and memory}

Skills and memory are adapted by changing their audience. The student-side skill states the workflow as direct instructions. Its adaptation in $\mathcal K$ tells the harnessing agent how to recognize a missing step and realize that step as a replacement. The following condensed excerpts illustrate the change:

\begin{lstlisting}[style=hzlisting]
Student-side skill:
  Read pre-filled examples before editing.
  Infer the rule from them and preserve their values.
  Before finishing, compare the output with the input.

Reference-harness guidance:
  Check whether the visible trajectory inspected the examples.
  If not, replace the current proposal with a bounded inspection.
  Before accepting a finish, require an input-output comparison.
\end{lstlisting}
\captionof{skilllisting}{Condensed skill adaptation for SpreadsheetBench.}\label{lst:skill-adaptation}

Memory is adapted in the same way, but starts from failures observed across training tasks. For example, the student-side memory records that agents often overwrite worked examples or verify formulas only against their own outputs. In $\mathcal K$, these observations become review-time patterns with concrete interventions:

\begin{lstlisting}[style=hzlisting]
Observed failure:
  The student overwrites pre-filled examples with its inferred rule.
Review condition:
  An edit spans the answer range without preserving existing literals.
Intervention:
  Replace with inspection or restore-and-diff actions.

Observed failure:
  Verification only rereads values produced by the same script.
Review condition:
  No assertion uses worked examples or independent recomputation.
Intervention:
  Replace with an assertion-based verification action.
\end{lstlisting}
\captionof{memorylisting}{Condensed memory adaptation for SpreadsheetBench.}\label{lst:memory-adaptation}

Across these adaptations, $\mathcal K$ retains the domain-specific condition and corrective behavior.
The harnessing agent supplies the final response, and $h$ remains the only interface through which that response acts on the environment.

\section{Review protocol and harnessing-agent prompt}\label{app:review-protocol}

This section details the interface between the student and the harnessing agent, and gives the system prompt used during training-data collection.

\subsection{Session structure}

One harnessing-agent session covers one complete student trial. The session is incremental: the harnessing agent keeps its full history across reviews, so earlier updates, its own earlier decisions, and any component files it has read remain in context.

Each review begins with a \emph{student update} message containing two parts. The first is the set of student-visible events added since the previous review, which on a typical turn is the observation produced by executing the previous accepted response. The second is the current unexecuted proposal, rendered as its reasoning, its visible content, and its tool call, or marked as a final response when it carries no tool call. The first update of a session additionally carries the student system prompt and the task instruction.

In addition, if the student emitted a structurally invalid response, such as more than one tool call (invalid for mini-SWE-agent) or a response with neither content nor a tool call, the update names the defect and states that passing it would preserve the defect. If a review middleware from $\mathcal K$ matches the current proposal, the update carries a triggered-guidance block with the middleware's evidence and hint.

\subsection{Submission format}

The harnessing agent may read adapted harness components under \texttt{/components} as many times as it needs before deciding, but it must submit exactly one decision per update by calling \texttt{submit\_review}. The submission has four fields the agent controls:

\begin{itemize}
\item \texttt{decision}, either \textsc{pass} or \textsc{replace};
\item \texttt{replacement}, the complete student response to execute instead of the proposal;
\item \texttt{components\_used}, the identifiers of the components that informed the decision;
\item \texttt{reason}, a private justification of at most 500 characters.
\end{itemize}

A replacement is a structured object with three fields: \texttt{reasoning}, \texttt{content}, and an optional \texttt{tool\_call}. For mini-SWE-agent, the only admissible tool call is a single \texttt{execute} with a non-empty command, which matches the student's action space under $h$ exactly. A replacement with no tool call and non-empty content is a final answer and ends the trial.

The schema rejects three malformed harnessing submissions: a \textsc{pass} that carries a replacement, a \textsc{replace} that does not, and a replacement with neither visible content nor a tool call. A submission that fails validation is not recorded. When an update produces no valid submission, the runtime re-prompts the harnessing agent for the same candidate, at most three times, and the trial fails if no valid submission is recorded by then.
Within one trial, the number of \textsc{replace} decisions the harnessing agent may issue is a hyperparameter. The budget is stated in the harnessing agent's system prompt, so that it can allocate interventions across the trial. We set it to 5, 3, and 1 to collect trajectories for SpreadsheetBench and AppWorld, and set it to 5 consistently for USPTO.

\subsection{System prompt}

\Cref{lst:teacher-prompt} gives the system prompt used to collect training data. Two variants exist. The training-free evaluations in \Cref{sec:exp-where} use a prompt with the same session structure, submission format, and budget, differing only in framing: it describes the goal as turning the components into better student actions rather than as producing a training trajectory, and it omits the paragraph on instructions regarding remaining on-policy. The oracle-answer control in \Cref{sec:exp-supervision} appends one extra section granting read access to the recorded reference answer for the current task. It also instructs the harnessing agent to use it only to decide whether the student's direction can still converge, to keep every verification step in the replacement, and never to reveal that the answer is known.

\begin{lstlisting}[style=hzlisting]
You are a strong harnessing agent supervising a student that solves a task with one `execute` tool and optional subagent delegation through `agent "<task>"` in bash.

The accepted trajectory of this trial will be used directly as SFT training data for the student model. Produce a successful trajectory that stays close to the student's behavior while using the teacher-side harness to introduce reusable task-solving patterns through targeted interventions. The student should do most of the work. When intervention is needed, replace the next response rather than taking over the task, then return control to the student after that response is executed.

On-policy here means that reviews occur along a live student trial and every accepted response uses the student's visible information, native action space, and plausible level of complexity. It does not mean preserving the student's current policy unchanged: a replacement should teach a better behavioral pattern when the harness identifies one that materially improves correctness, progress, recovery, or verification.

## Session input

One teacher session covers one complete student trial. Each `Student update` message contains the student-visible events added since the previous review and the current unexecuted proposal. The first update also contains the student system prompt and task.

The student sandbox is not mounted in your workspace. Treat file contents, command results, installed programs, and service state as known only when they appear in a student-visible observation.

## Components

Read `/components/index.json` before the first decision. Choose and read memory sections or component files as they become relevant during the trial.

Mounted middleware may append `Triggered middleware guidance` to a student update when the current unexecuted proposal matches one of its checks. Use the stated evidence and hint when reviewing that proposal.

Use the components actively as your knowledge of what good behavior looks like. They should affect the accepted trajectory when their guidance is relevant, not merely help you recognize fatal errors. Procedural knowledge from a component may inform a replacement even if the student has not demonstrated it yet, provided the resulting response is a plausible next step in the student's native interface. Component-private facts, paths, review records, and unsupported claims about the current sandbox must never enter the replacement.

## Review

Review every proposed tool call and final answer before it is accepted.

Choose `PASS` when the proposal is a sound next action and no applicable component calls for a meaningful behavioral correction. Pass harmless inefficiencies, stylistic differences, valid alternative methods, and exploratory steps that can produce useful evidence. Also pass recoverable mistakes when observing the result is likely to let the student diagnose and repair them; useful self-recovery is valuable training behavior.

Choose `REPLACE` when a meaningful correction is needed for task success or to instantiate a reusable pattern supplied by the harness. Common reasons include:

- a missed requirement, damaged or skipped deliverable, unsupported conclusion, or premature final answer;
- an unsafe, unbounded, fragile, or budget-wasting command;
- an observed failure that the student ignores or repeats without a useful change;
- a missing inspection, dependency check, test, or verification step needed to ground later work;
- an applicable component identifies a behavior pattern that the proposal violates or omits, and correcting it now would materially improve progress, recoverability, or the value of the trajectory.

You may replace at most `5` student responses during the complete trial. The budget rewards selectivity, not passivity: do not polish sound actions, but do not withhold a useful pattern-level correction merely because the current proposal is not immediately fatal.

When you replace:

- Make the smallest coherent change to the student's next step that installs the correction. This may require a different action, not merely a textual patch.
- Preserve the student's high-level intent, established facts, language, variable names, and command structure when they remain compatible with the correction.
- Stay near the student's demonstrated level, but you may introduce a simple procedure, command, library, or idiom from the components when it is needed to express the target pattern. Make the reasoning understandable from student-visible evidence rather than relying on unexplained teacher expertise.
- Correct the next decision and hand control back. Do not complete several future steps, precompute the task's answer, or replace work the student can perform after seeing the next observation.
- The reasoning must read as the student's own first-person inner monologue, continuing the student's current line of thought -- never as advice, critique, or correction addressed to the student. A natural form is the student catching its own mistake: noticing the constraint, re-reading the evidence, and correcting course.
- A replacement must contain complete student-style reasoning, visible content, and at most one `execute` call, or a final answer without a tool call. Its reasoning, content, and tool call must agree. Base every factual claim on student-visible events.

Do not mention the teacher, review, harness components, component names, private paths, or training in the replacement.

Pass a final answer only after the student-visible observations support every material task requirement. Otherwise replace it with the next inspection, repair, test, or verification action, not with a teacher-written solution that bypasses those steps.

## Submission

Call `submit_review` exactly once for each student update. For `PASS`, set `replacement` to null. For `REPLACE`, provide the complete replacement. Record the components you used and a concise reason in the private submission metadata.
\end{lstlisting}
\captionof{promptlisting}{System prompt of the harnessing agent during training-data collection.}\label{lst:teacher-prompt}

\section{Trajectory collection and filtering}\label{app:data-collection}

On SpreadsheetBench and AppWorld, we retain a trial only if its verifier reward is 1.0 and the run finished without an execution exception. USPTO keeps every collected trial, so the ablation in \Cref{sec:exp-supervision} trains each condition on the same 500 tasks. Training on this full USPTO set also yields higher test performance than keeping successes only.

Every training trajectory must contain only student-visible text. We scan each trajectory for strings that can appear only in the private review and filter out any trajectory that contains the following.

\begin{itemize}
\item an absolute \texttt{/components/} path, which can only refer to the reference harness mount;
\item the path of the recorded candidate file that stores the unexecuted proposal;
\item the name of the review submission tool;
\item the review metadata key that records which components a decision used;
\item the internal name of the review middleware.
\end{itemize}

A replacement is written inside the review context, so its reasoning can slip into the reviewer's voice and describe the student's proposal from the outside. 
We mask the reasoning token span of any accepted response that was produced by a \textsc{replace} decision and whose reasoning matches a reviewer-perspective pattern.
The patterns cover the words \emph{proposal} and \emph{proposed}, references to a draft, references to the student's or the candidate's response, reasoning, command, or action, and explicit review verbs applied to a response, such as passing or rewriting it. The resulting training sets contain 487 trajectories for SpreadsheetBench, 282 for AppWorld, and 500 for USPTO, as reported in \Cref{sec:exp-setup}.

\begin{table}[!ht]
\centering
\hztablesetup
\setlength{\tabcolsep}{2.0pt}
\caption{Training-set composition and filtering statistics across the three domains.}
\label{tab:training-set-stats}
\begin{tabular}{lccc}
\hztablerule
\rowcolor{hzheader}
\textbf{Metric} & \textbf{SpreadsheetBench} & \textbf{AppWorld} & \textbf{USPTO} \\
\hztablerule
Collection trials per task & 3 / 2 (700 total) & 3 (441 total) & 1 (500 total) \\
Collection success rate ($\text{reward}=1.0$) & 81.3\% (569/700) & 66.0\% (291/441) & 59.4\% (297/500) \\
Retained SFT trajectories & 487 & 282 & 500 \\
Distinct training tasks covered & 264/300 (88.0\%) & 133/147 (90.5\%) & 500/500 (100.0\%) \\
\textsc{replace} turn fraction & 53.6\% (1,539/2,869) & 14.9\% (794/5,336) & 88.4\% (1,749/1,978) \\
Teacher-written token fraction & 49.5\% & 13.4\% & 94.1\% \\
Masked reasoning spans (\% of \textsc{replace}) & 171 (11.1\%) & 21 (2.6\%) & 143 (8.2\%) \\
Dropped by privacy filter / 65,536-token limit & 0 / 1 & 0 / 0 & 0 / 0 \\
\hztablerule
\end{tabular}
\end{table}

\paragraph{Training-set composition and trajectory audit.}
\Cref{tab:training-set-stats} reports detailed collection and filtering statistics for each domain.
Notably, the fraction of teacher-written tokens does not explain whether distillation surpasses $h^\star$. On AppWorld, where \our{} exceeds $h^\star$ by the largest margin (+10.7 points), 85.1\% of accepted turns are \textsc{pass} turns written by the student itself (only 13.4\% teacher-written tokens), whereas on USPTO, where the harnessing agent replaces 88.4\% of turns, the distilled student does not surpass $h^\star$.
Across all 4,082 \textsc{replace} steps in the three training sets, deterministic checks found 0\% leakage of private paths or component names and masked the SFT loss on the 8.2\% of steps matching reviewer-perspective phrasing while keeping their tool calls trainable.
On a random sample of 100 remaining unmasked \textsc{replace} steps, human and LLM-judge audits found no third-person reviewer voice or ungrounded task premises. Only 1.5\% of audit evaluations flagged reasoning text that briefly referenced the student's draft command.

\begin{table}[!ht]
\centering
\hztablesetup
\setlength{\tabcolsep}{2.0pt}
\caption{Comparing alternative distillation supervision sources across the three benchmarks (all evaluated under $h$). The last two columns report the fraction of test trajectories that repeat an identical command at least five times.}
\label{tab:matched-controls}
\begin{tabular}{lccccc}
\hztablerule
\rowcolor{hzheader}
\textbf{Supervision Source (under $h$)} & \multicolumn{3}{c}{\textbf{Task Success (\%)}} & \multicolumn{2}{c}{\textbf{Repeated Cmd.\ ($\ge 5\times$, \%)}} \\
\rowcolor{hzheader}
& SpreadsheetBench & AppWorld & USPTO & SpreadsheetBench & AppWorld \\
\hztablerule
Base model (untrained) & 31.0 & 26.8 & 12.0 & --- & --- \\
Rejection-sampling SFT ($2\times$ budget) & 31.0 & \snd{50.0} & \snd{14.0} & --- & --- \\
Standalone teacher rollouts & 20.0 & 32.1 & 12.0 & 13.0 & 20.8 \\
Review with empty $\mathcal K$ & 28.0 & 48.2 & 11.0 & 4.0 & 4.2 \\
\textbf{\our{} ($\mathcal K$, unfiltered)} & \snd{43.0} & --- & \textbf{30.0} & --- & --- \\
\textbf{\our{} ($\mathcal K$)} & \textbf{44.0} & \textbf{58.9} & \textbf{30.0} & \textbf{0.0} & \textbf{1.2} \\
\hztablerule
\end{tabular}
\end{table}

\paragraph{Further ablations on SpreadsheetBench and AppWorld.}
\Cref{tab:matched-controls} extends the supervision-source ablations (\Cref{sec:exp-supervision}) to SpreadsheetBench and AppWorld under the same SFT recipe, showing that \our{} consistently outperforms alternative supervision sources across all three domains.
First, training on all 691 \emph{unfiltered \our{}} trajectories on SpreadsheetBench without reward filtering reaches 43.0\%, nearly matching the filtered set (44.0\%) and surpassing $h^\star$ (39.0\%); this confirms that our distillation gains do not depend on verifier filtering.
Second, \emph{rejection-sampling SFT (RFT)} trails \our{} by 13.0, 8.9, and 16.0 points across the three domains even when given $2\times$ the rollout budget (1,800 rollouts on SpreadsheetBench, 882 on AppWorld, and 1,000 on USPTO) and 46\%--76\% more successful training trajectories (713 vs.\ 487 on SpreadsheetBench and 497 vs.\ 282 on AppWorld), as self-sampling mostly reinforces tasks the student can already solve.
Third, when matched to \our{} in training set size (487 trajectories on SpreadsheetBench and 282 on AppWorld), \emph{standalone teacher rollouts} reach only 20.0\% and 32.1\% on the two benchmarks, and \emph{empty-$\mathcal K$ review} reaches 28.0\% and 48.2\%, trailing \our{} by 22.9 and 15.2 points in three-domain macro-average.
Finally, the last two columns of \Cref{tab:matched-controls} show that students distilled from standalone off-policy teacher rollouts frequently get stuck repeating an identical failing command at least five times at test time (13.0\% on SpreadsheetBench and 20.8\% on AppWorld), whereas on-policy, $\mathcal K$-guided step-level corrections largely prevent such loops (0.0\% and 1.2\% for \our{}).

\section{Implementation details}\label{app:implementation}

\subsection{Student harness}

The system prompt of mini-SWE-agent $h$ is given in \Cref{lst:student-prompt}. It is the only instruction the student receives beyond the task itself.

\begin{lstlisting}[style=hzlisting]
You are an agent that solves tasks in a Linux sandbox.

You have one tool, `execute`, which runs one bash command. Each call starts a fresh shell, so working-directory and environment changes do not persist between calls.

Work on the task by issuing one `execute` call at a time. When the task is complete, return a short final answer without calling `execute`.

The sandbox also provides `agent "<task>"` for optional subagent delegation.
\end{lstlisting}
\captionof{promptlisting}{System prompt of the student under the target harness $h$.}\label{lst:student-prompt}

The harness exposes exactly one tool \texttt{execute}, whose single argument is a non-empty bash command. Each call runs in a fresh shell, so the working directory and environment variables do not carry over; state persists only through the file system. The observation returned to the student is the command's combined output followed by a line giving its exit code. A response with no tool call ends the trial and is taken as the final answer.

The sandbox additionally provides an \texttt{agent "<task>"} CLI utility for single-level subagent delegation. A subagent runs the same base model with the single \texttt{execute} tool, receives a fixed prompt instructing it to return a concise report for its bounded task, and cannot delegate further. At most four subagents may run concurrently per sandbox. This capability ensures that bash plus one-level delegation forms an action space expressive enough to map diverse evolved $h^\star$ components onto $h$. In practice, the evolved $h^\star$ in this work do not rely on subagents, and delegation is used sparingly.

\subsection{Models and access}

We access GPT-5.6 Sol~\citep{openai2026gpt56}, DeepSeek-V4-Pro, and DeepSeek-V4-Flash~\citep{deepseekai2026deepseekv4} through Microsoft Azure. Qwen3.5-9B~\citep{qwen2026qwen35} is evaluated as the base model through OpenRouter~\citep{openrouter2025}, as is Qwen3.6-35B-A3B in the analysis of \Cref{app:aah-analysis}, and the fine-tuned checkpoints are served on NVIDIA A100 GPUs. All harness-evolution runs are performed by Kimi K3~\citep{moonshot2026kimik3} using Kimi Code~\citep{moonshot2026kimicode}. We use the default decoding parameters provided by each model provider. Reasoning is enabled for all models, with reasoning effort set to \emph{high} when applicable.

\subsection{Training configuration}

We perform LoRA supervised fine-tuning on Qwen3.5-9B with the Tinker supervised-training recipe~\citep{thinkingmachines2025tinker}, using rank 32, $\alpha{=}32$, batch size 8, a 65{,}536-token sequence length, and two epochs. The learning rate follows a linear schedule with 5\% warmup to a peak of $2\times10^{-4}$, followed by decay to $10^{-6}$. All main runs use seed 42, and examples longer than the sequence length are dropped.
To verify training stability across random seeds, we additionally train \our{} on SpreadsheetBench with seeds 43 and 44, achieving 43.0\% and 49.0\% test pass@1 (alongside 44.0\% for seed 42), consistently outperforming both the untrained base model (31.0\%) and the base model with $h^\star$ attached (39.0\%).

\section{Additional analysis of agent-as-harness}\label{app:aah-analysis}

We examine whether the advantage of agent-as-harness extends to weaker models when the same model serves as both the student and the harnessing agent.
\Cref{tab:aah-weaker} extends the USPTO comparison of \Cref{tab:train-free} to two additional, weaker models.

\begin{table}[!ht]
\centering
\hztablesetup
\caption{USPTO pass@1 (\%) on two weaker models, using the four conditions of \Cref{tab:train-free}. Together with the USPTO rows of \Cref{tab:train-free}, the models span four capability tiers in decreasing order: GPT-5.6 Sol, DeepSeek-V4-Pro, DeepSeek-V4-Flash, Qwen3.6-35B-A3B.}
\label{tab:aah-weaker}
\begin{tabular}{lrrrr}
\hztablerule
\rowcolor{hzheader}
& \multicolumn{2}{c}{\textbf{Code-as-Harness}} & \multicolumn{2}{c}{\textbf{Agent-as-Harness}} \\
\rowcolor{hzheader}
\multicolumn{1}{c}{\textbf{Model}} & \makecell{mini-SWE-agent\\($h$)} & \makecell{meta-harness\\($h^\star$)} & \makecell{w/o evolved\\($h,\varnothing$)} & \makecell{w/ evolved\\($h,\mathcal K$)} \\
\hztablerule
DeepSeek-V4-Flash  & 33.0 & \textbf{58.0} & 32.3 & 57.0 \\
Qwen3.6-35B-A3B    & 16.0 & \textbf{48.0} & 12.0 & 36.0 \\
\hztablerule
\end{tabular}
\end{table}

The relative advantage of agent-as-harness over meta-harness ($(h, \mathcal K) - h^\star$) closely correlates with model capability. It remains positive on stronger models but turns negative on weaker ones (+1.0\% on GPT-5.6 Sol, +4.0\% on DeepSeek-V4-Pro, -1.0\% on DeepSeek-V4-Flash, and -12.0\% on Qwen3.6-35B-A3B).
On the weakest model, the harnessing agent intervenes aggressively, replacing 66.0\% of reviewed steps, but its edits are net harmful. Even review without evolved guidance falls below mini-SWE-agent (12.0 vs.\ 16.0).
Effective review and intervention therefore require sufficient underlying model capability.

The current \our{} design reviews every student proposal before execution, requiring at least two model calls per step: one to generate the proposal and another to review it.
Each review processes the trajectory prefix and reference harness, and must finish before the next student step, which increases both token use and latency.
On USPTO, agent-as-harness with evolved guidance takes 237.2\,s per trial on average, about $2.4\times$ the 100.1\,s required by mini-SWE-agent.
Reducing the overhead, for example by reviewing only a subset of steps, is left to future work.

\section{Harness-exclusive behavior patterns}\label{app:behavior-patterns}

Here we detail how the behavior patterns of \Cref{sec:exp-behavior} are mined and explain the mined patterns on the three benchmarks.
All measurements use the trajectories of three deployments from \Cref{tab:distillation}: mini-SWE-agent ($h$), meta-harness ($h^\star$), and \our{} ($h$).

\subsection{Approach}

For each domain we start from the enabled components of the student-side harness $h^\star$ and read every enabled memory file, skill file, tool, and middleware. Then, we translate each rule into a deterministic binary detector over the trajectory (agent messages, bash commands, and observations).
Detectors match the target behavior semantically.
For example, AppWorld pagination counts if the trajectory uses the \texttt{aw.pages} helper or an explicit loop over \texttt{page\_index}; a USPTO answer write counts if it uses a shell redirect or Python \texttt{write\_text}.
The USPTO tools \texttt{smiles\_check}, \texttt{propose\_retrosynthesis}, and \texttt{verify\_answer} exist only under $h^\star$. A call to them counts on $h^\star$ trajectories, and an equivalent bash or Python command counts under $h$.

For each detector, we define the support tasks as the test tasks on which, for the base model, the pattern appears under $h^\star$ but not under $h$.
We keep a detector only if it has at least 10 support tasks.
If two detectors give the same result on every applicable task, they are measuring the same behavior, so we merge them into one pattern.
This procedure yields 18 patterns on SpreadsheetBench, 6 on USPTO, and 4 on AppWorld.
A pattern's \emph{recovery rate} is the fraction of its support tasks on which \our{} exhibits the behavior. By construction, the base model under $h$ and $h^\star$ exhibits it on 0\% and 100\% of these tasks, respectively.

\subsection{Mined patterns}

\Cref{tab:app-patterns-ssb,tab:app-patterns-uspto,tab:app-patterns-appworld} list every retained pattern.
\begin{table}[!ht]
\centering
\hztablesetup
\setlength{\tabcolsep}{4pt}
\caption{4 harness-exclusive AppWorld patterns.}
\label{tab:app-patterns-appworld}
\begin{tabular}{>{\raggedright\arraybackslash}p{0.18\linewidth}>{\raggedright\arraybackslash}p{0.20\linewidth}>{\raggedright\arraybackslash}p{0.36\linewidth}r}
\hztablerule
\rowcolor{hzheader}
\multicolumn{1}{c}{\textbf{Pattern}} & \multicolumn{1}{c}{\textbf{Source}} & \multicolumn{1}{c}{\textbf{Mined rule}} & \multicolumn{1}{c}{\textbf{Recovery}} \\
\hztablerule
Retrieve complete pages & aw.py & Fetch until a terminal empty page; never rely on the implicit first page. & 21/53 (40\%) \\
Read back mutations & skill & Read back affected state after a mutation. & 6/12 (50\%) \\
Avoid repeated failed calls & \path{failure_recovery_instruction} & Do not repeatedly retry an unchanged failed call. & 22/31 (71\%) \\
Choose action vs.\ answer & submit\_gate & Use an answer only for information requests; none for actions. & 24/33 (73\%) \\
\hztablerule
\end{tabular}
\end{table}

\begin{table}[!ht]
\centering
\hztablesetup
\setlength{\tabcolsep}{4pt}
\caption{18 harness-exclusive SpreadsheetBench patterns. We gather the harness source, the mined rule (abbreviated), the support size, and \our{} recovery.}
\label{tab:app-patterns-ssb}
\begin{tabular}{>{\raggedright\arraybackslash}p{0.18\linewidth}>{\raggedright\arraybackslash}p{0.20\linewidth}>{\raggedright\arraybackslash}p{0.36\linewidth}r}
\hztablerule
\rowcolor{hzheader}
\multicolumn{1}{c}{\textbf{Pattern}} & \multicolumn{1}{c}{\textbf{Source}} & \multicolumn{1}{c}{\textbf{Mined rule}} & \multicolumn{1}{c}{\textbf{Recovery}} \\
\hztablerule
Edit existing workbook & memory.md & Deliver the input workbook edited in place; never rebuild it. & 31/34 (91\%) \\
Assert workbook structure & memory.md & Reload and assert sheet names and headers match the input. & 20/27 (74\%) \\
Protect pre-filled cells & memory.md + \path{prefilled_guard} & Read and preserve pre-filled cells; write only blank cells. & 24/26 (92\%) \\
Write computed literals & memory.md + \path{answer_range_guard} & Write literal values, not unverifiable formula strings. & 19/21 (90\%) \\
Avoid fragile coordinates & memory.md & Use Excel coordinates; never \texttt{chr(64+n)} or positional writes. & 19/19 (100\%) \\
Save requested workbook & memory.md + output\_guard & The deliverable is the manipulated workbook at the requested path. & 14/14 (100\%) \\
Inspect before editing & skill & Explore sheets, headers, dimensions, and examples first. & 32/32 (100\%) \\
Inspect target/examples & skill + \path{prefilled_guard} & Inspect the requested output area and examples before writing. & 14/18 (78\%) \\
Use \texttt{solution.py} & skill & Write \texttt{solution.py} with input and output paths at the top. & 53/76 (70\%) \\
Execute final script & skill & Execute \texttt{python solution.py} before finishing. & 52/75 (69\%) \\
Read cached formulas & skill & Use \texttt{data\_only=True} when existing formulas are inputs. & 28/43 (65\%) \\
Normalize matching keys & skill & Normalize text, numeric, and date keys deliberately for matching. & 6/10 (60\%) \\
Use closed target ranges & skill & Treat requested ranges as closed intervals and honor their bounds. & 30/32 (94\%) \\
Reload saved workbook & skill & Reload the saved workbook and verify representative target cells. & 17/17 (100\%) \\
Check output existence & output\_guard & Check that the output workbook exists and is non-empty. & 16/16 (100\%) \\
Populate answer range & answer\_range\_guard & Do not finish with an all-empty answer range. & 28/30 (93\%) \\
Scan range after save & answer\_range\_guard & Reload and mechanically inspect answer-range cells before finishing. & 20/24 (83\%) \\
Stop after delivery & turn\_budget & Stop no-op verification once a valid deliverable is saved. & 21/28 (75\%) \\
\hztablerule
\end{tabular}
\end{table}

\begin{table}[!ht]
\centering
\hztablesetup
\setlength{\tabcolsep}{4pt}
\caption{6 harness-exclusive USPTO patterns.}
\label{tab:app-patterns-uspto}
\begin{tabular}{>{\raggedright\arraybackslash}p{0.18\linewidth}>{\raggedright\arraybackslash}p{0.20\linewidth}>{\raggedright\arraybackslash}p{0.36\linewidth}r}
\hztablerule
\rowcolor{hzheader}
\multicolumn{1}{c}{\textbf{Pattern}} & \multicolumn{1}{c}{\textbf{Source}} & \multicolumn{1}{c}{\textbf{Mined rule}} & \multicolumn{1}{c}{\textbf{Recovery}} \\
\hztablerule
Parse product with RDKit first & skill & Parse the product with RDKit first; never reason from the raw string. & 23/23 (100\%) \\
Enumerate candidate sets & skill + \path{propose_retrosynthesis} & Enumerate at least 3 candidate reactant sets before committing. & 59/88 (67\%) \\
Validate with RDKit & memory.md + \path{smiles_check} & Never eyeball-parse a SMILES; validate every structure with RDKit. & 60/60 (100\%) \\
Canonicalize SMILES & memory.md + \path{smiles_check} & Ship every reactant in RDKit-canonical form. & 73/73 (100\%) \\
Check heavy-atom coverage & memory.md + \path{verify_answer} & Reactant heavy atoms must account for every product heavy atom. & 40/54 (74\%) \\
Validate the final answer & memory.md + \path{answer_guard} & Validation must cover the answer file's last write. & 45/47 (96\%) \\
\hztablerule
\end{tabular}
\end{table}

\end{document}